\documentclass[acmtog]{acmart} 

\usepackage{booktabs} 

\usepackage[ruled]{algorithm2e} 

\SetAlFnt{\small}
\SetAlCapFnt{\small}
\SetAlCapNameFnt{\small}
\SetAlCapHSkip{0pt}

\acmJournal{TOG}

\copyrightyear{2024} 
\acmYear{2024} 
\setcopyright{acmlicensed}\acmConference[SA Conference Papers '24]{SIGGRAPH Asia 2024 Conference Papers}{December 3--6, 2024}{Tokyo, Japan}
\acmBooktitle{SIGGRAPH Asia 2024 Conference Papers (SA Conference Papers '24), December 3--6, 2024, Tokyo, Japan}
\acmDOI{10.1145/3680528.3687658}
\acmISBN{979-8-4007-1131-2/24/12}

\newcommand{\RI}{\textit{Relation Inversion}} 
\newcommand{\R}{$\langle$R$\rangle$} 

\usepackage{caption}

\usepackage{booktabs, siunitx}
\usepackage{multirow, booktabs}
\usepackage{siunitx}
\usepackage{makecell} 

\usepackage{mathtools}

\newcommand{\defeq}{\coloneqq}

\newcommand{\E}{\mathbb{E}}

\newcommand{\Eb}[2]{\E_{#1}\!\left[#2\right]}

\newcommand{\bI}{\mathbf{I}}

\newcommand{\bzero}{\mathbf{0}}

\newcommand{\bx}{\mathbf{x}}

\newcommand{\bepsilon}{{\boldsymbol{\epsilon}}}

\usepackage{amsmath}

\DeclareMathOperator*{\argmin}{arg\,min}

\usepackage{marvosym, ifsym}

\usepackage{multicol}

\usepackage{mainstyle}

\usepackage{subcaption} 

\newcommand{\extension}[1]{{\color{black}{#1}}} 

\begin{document}
\title{ReVersion: Diffusion-Based Relation Inversion from Images}

\author{Ziqi Huang}
\authornotemark[1]
\email{ziqi002@ntu.edu.sg}
\orcid{0000-0001-8008-5873}
\affiliation{%
\institution{S-Lab, Nanyang Technological University}
\country{Singapore}
}

\author{Tianxing Wu}
    \authornote{Equal contributions}
    \email{tianxing001@ntu.edu.sg}
    \orcid{0000-0001-7345-0254}
    \affiliation{%
    \institution{S-Lab, Nanyang Technological University}
    \country{Singapore}
}
\author{Yuming Jiang}
    \email{yuming002@ntu.edu.sg}
    \orcid{0000-0001-7653-4015}
    \affiliation{%
    \institution{S-Lab, Nanyang Technological University}
    \country{Singapore}
}
\author{Kelvin C.K. Chan}
    \email{chan0899@ntu.edu.sg}
    \orcid{0000-0002-5456-8991}
    \affiliation{%
    \institution{S-Lab, Nanyang Technological University}
    \country{Singapore}
}
\author{Ziwei Liu}
    \authornote{Corresponding author}
    \email{ziwei.liu@ntu.edu.sg}
    \orcid{0000-0002-4220-5958}
    \affiliation{%
    \institution{S-Lab, Nanyang Technological University}
    \country{Singapore}
}

\begin{abstract}
    Diffusion models gain increasing popularity for their generative capabilities. Recently, there have been surging needs to generate customized images by inverting diffusion models from exemplar images, and existing inversion methods mainly focus on capturing object \textbf{appearances} (\ie, the ``look''). However, how to invert object \textbf{relations}, another important pillar in the visual world, remains unexplored.
    In this work, we propose the \textbf{\RI{}} task, which aims to learn a specific relation (represented as ``relation prompt'') from exemplar images. Specifically, we learn a relation prompt with a frozen pre-trained text-to-image diffusion model. The learned relation prompt can then be applied to generate relation-specific images with new objects, backgrounds, and styles. 
    To tackle the \RI{} task, we propose the \textbf{\textit{ReVersion Framework}}.
    Specifically, we propose a novel ``relation-steering contrastive learning'' scheme to steer the relation prompt towards relation-dense regions, and disentangle it away from object appearances. 
    We further devise ``relation-focal importance sampling'' to emphasize high-level interactions over low-level appearances (\eg, texture, color).
    To comprehensively evaluate this new task, we contribute the \textbf{ReVersion Benchmark}, which provides various exemplar images with diverse relations. Extensive experiments validate the superiority of our approach over existing methods across a wide range of visual relations. Our proposed task and method could be good inspirations for future research in various domains like generative inversion, few-shot learning, and visual relation detection.
    
\end{abstract}



\begin{teaserfigure}
    \includegraphics[width=\textwidth]{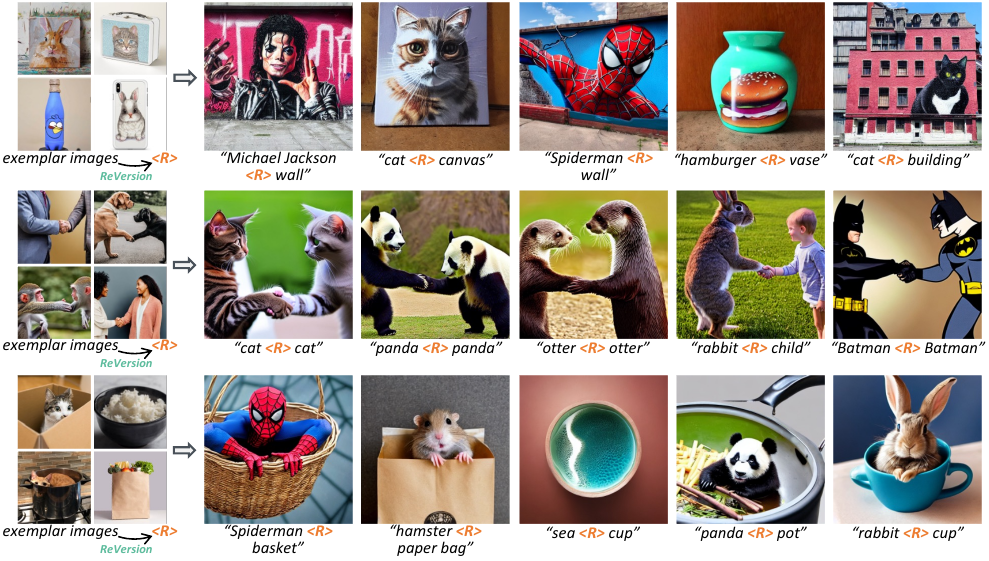}
    \vspace{-20pt}
    \caption{
        We propose a new task, \textbf{Relation Inversion}: Given a few exemplar images, where a relation co-exists in every image, we aim to find a relation prompt \R{} to capture this interaction, and apply the relation to new entities to synthesize new scenes. The above images are generated by our \textbf{ReVersion Framework}.
    }
        \label{fig:teaser}
\end{teaserfigure}

\maketitle

\vspace{15pt}
\section{Introduction}
\label{sec:intro}

Recently, text-to-image (T2I) diffusion models~\cite{rombach2022ldm, ramesh2022dalle2, saharia2022imagen} have shown promising results and enabled subsequent explorations of various generative tasks. 
There have been several explorations~\cite{gal2022textualinversion, ruiz2022dreambooth, kumari2022customdiffusion, chen2023suti, wei2023elite, li2023blipdiffusion, jia2023tamingencoder} on the \textit{appearance inversion} task. Specifically, given a few images of a specific object (\eg, a cat statue), \textit{appearance inversion} learns to map a ``new word'' to this concept via the text-to-image model. 
The ``new word'' can then be used in prompts to generate new images that contain this concept.
While existing methods have made substantial progress in capturing object \textit{appearances}, such exploration for \textit{relations} between objects is rare. 

In this paper, we study the \textbf{\textit{Relation Inversion}} task, whose objective is to learn a \textit{relation} that co-exists in the given exemplar images. Specifically, with objects in each exemplar image following a specific relation, we aim to obtain a relation prompt in the text embedding space of the pre-trained text-to-image diffusion model. By composing the relation prompt with user-devised text prompts, users are able to synthesize images using the corresponding relation, with new objects, styles, and backgrounds, \etc. 
Studying \RI{} not only addresses a critical gap in text-to-image model inversion tasks but also paves the way for deeper understanding and generation of relation-rich visual content.

The \textit{Relation Inversion} task is intrinsically different from existing appearance inversion tasks, and thus poses unique challenges.
Appearance inversion~\cite{gal2022textualinversion, ruiz2022dreambooth, kumari2022customdiffusion, chen2023suti, wei2023elite, li2023blipdiffusion, jia2023tamingencoder} focuses on capturing the look of a specific entity, thus the commonly used pixel-level reconstruction loss is typically adequate to learn a prompt that encapsulates the shared information among exemplar images. In contrast, \textit{relation} is a more abstract visual concept, and a pixel-wise loss alone is insufficient for precise extraction of the intended relation. 
Consequently, linguistic and visual priors are needed to accurately represent these high-level relation concepts.

\begin{figure}[t]
    \vspace{6pt}
    \centering
     \includegraphics[width=0.7\linewidth]{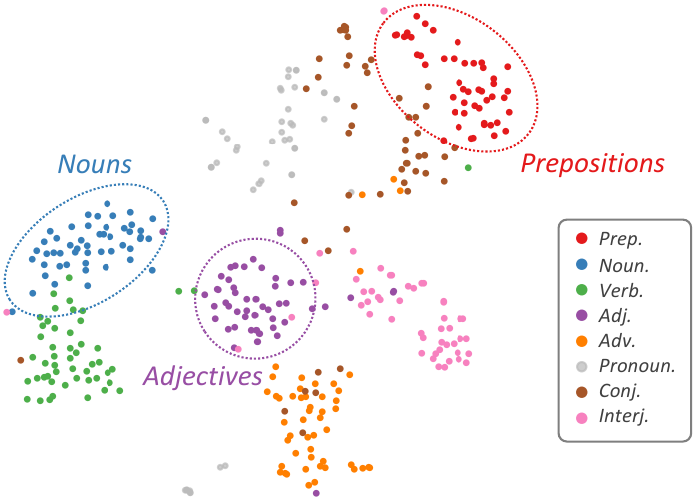}
     \caption{
        \textbf{Part-of-Speech (POS) Clustering}. We use t-SNE~\cite{van2008visualizing} to visualize word distribution in CLIP's input text embedding space, where \R{} is optimized in our ReVersion framework. We observe that words of the same Part-of-Speech (POS) are closely clustered together, while words of different POS are generally at a distance from each other.
    }
     \label{fig:observation_a}
  \end{figure}

To this end, we propose the \textit{\textbf{ReVersion Framework}} to tackle the \RI{} problem. 
First, we exploit linguistic priors to steer the relation prompt in the text embedding space. Specifically, we found that in the text embedding space of Stable Diffusion, embeddings are generally clustered according to their Part-of-Speech (POS), as shown in Figure~\ref{fig:observation_a}. Also, the concept of ``relation'' is related to prepositional words. For example, the relation ``rides on'' is semantically related to the prepositions ``atop'', ``above'', and ``below''; the relation ``being contained within'' is semantically related to ``inside'', ``around'', ``in'', and ``including''.
This together with the POS clustering observation motivate us to steer the relation prompts towards the prepositional word cluster. Notably, we design a novel \textit{relation-steering contrastive learning scheme} to steer the relation prompt towards a relatively relation-dense region in the text embedding space. A set of basis prepositions are used as positive samples to pull the relation prompt, while words of other POS (\eg, nouns, adjectives) in text descriptions are regarded as negatives so that the semantics related to object appearances are disentangled away.

Second, to encourage attention on object interactions, we devise a \textit{relation-focal importance sampling} strategy. During the optimization process, we emphasize high-level interactions over relatively lower-level details (\eg, color and texture of objects), effectively leading to better \RI{} results.

As the first attempt in this direction, we further contribute the \textit{\textbf{ReVersion Benchmark}}, which provides various exemplar images with diverse relations, from simple spatial arrangements to complex interactive behaviours.
The benchmark serves as an evaluation tool for future research in \textit{Relation Inversion}. Results on a variety of relations demonstrate the effectiveness of our ReVersion Framework.

Our contributions are summarized as follows:
\begin{itemize}
    \setlength\itemsep{0em}
    \item We study a new problem, \textbf{\textit{Relation Inversion}}, which requires learning a relation prompt for a relation that co-exists in several exemplar images. While existing T2I inversion works mainly focus on capturing appearances, we take the initiative to explore relation, an under-explored yet important pillar in the visual world.
    \item We propose the \textbf{\textit{ReVersion Framework}}, where the \textit{relation-steering contrastive learning scheme} steers relation prompt using linguistic priors, and effectively disentangles the learned relation away from object appearances. The \textit{relation-focal importance sampling} further emphasizes high-level relations over low-level details.
    \item We contribute the \textit{\textbf{ReVersion Benchmark}}, which serves as a diagnostic and benchmarking tool for the new task of \RI{}. 
\end{itemize}

\section{Related Work}

\begin{figure*}[t]
  \centering
  \vspace{6pt}
   \includegraphics[width=0.99\textwidth]{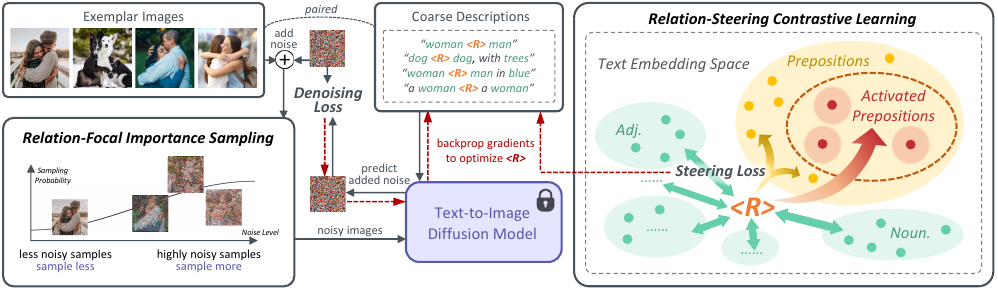}
   \caption{\textbf{ReVersion Framework}. Given exemplar images and their entities' coarse descriptions, our ReVersion framework optimizes the relation prompt \R{} to capture the relation that co-exists in all the exemplar images. During optmization, the \textit{relation-focal importance sampling} strategy encourages \R{} to focus on high-level relations, and the \textit{relation-steering contrastive learning} scheme induces the relation prompt \R{} towards relation-dense regions and away from entities or appearances. Upon optimization, \R{} can be used as a word in new sentences to make novel entities interact via the relation in exemplar images. 
   }
   \label{fig:framework}
   \vspace{6pt}
\end{figure*}

\noindent\textbf{Diffusion Models.}
Diffusion models~\cite{ho2020ddpm, sohl2015deep, song2020score, rombach2022ldm, gu2022vqdiffusion, song2020ddim} have become a mainstream approach for image synthesis~\cite{dhariwal2021beatgan, esser2021imagebart, meng2021sdedit} apart from GANs~\cite{goodfellow2014gan}, 
and have shown success in various domains such as video generation~\cite{harvey2022fdm,villegas2022phenaki,singer2022makeavideo,ho2022videoDM, he2022lvdm, wu2022tuneavideo, blattmann2023videoldm}, image restoration~\cite{saharia2022sr3, ho2022cascaded}, and many more~\cite{baranchuk2021label,graikos2022diffusion, amit2021segdiff, austin2021structured}.
Diffusion models are usually trained using score-matching objectives~\cite{hyvarinen2005estimation, vincent2011connection} at various noise levels, and sampling is done via iterative denoising. 
Text-to-Image (T2I) diffusion models~\cite{ramesh2022dalle2, rombach2022ldm, esser2021imagebart, gu2022vqdiffusion, jiang2022text2human, nichol2021glide, saharia2022imagen} demonstrated impressive results in converting user-provided text descriptions into images. 
Motivated by their success, we build our framework on a state-of-the-art T2I diffusion model, Stable Diffusion~\cite{rombach2022ldm}.

\noindent\textbf{Relation Modeling.}
Relation modeling has been explored in discriminative tasks such as scene graph generation~\cite{xu2017scene,vg17ijcv,shang2017video,ji2020action, yang2022psg, yang2023pvsg} and visual relationship detection~\cite{lu2016visual, yu2017visual, zhuang2017towards}.
These works aim to detect visual relations between objects in given images and classify them into a predefined, closed-set of relations.
However, the finite relation category set intrinsically limits the diversity of captured relations.
In contrast, \RI{} regards relation modeling as a generative task, aiming to capture arbitrary, open-world relations from exemplar images and apply the resulting relation for content creation.

\noindent\textbf{Diffusion-Based Inversion.} Given a pre-trained T2I diffusion model, \textit{inversion} aims to find a text embedding vector to express the concepts in the given exemplar images, via optimization-based~\extension{\cite{gal2022textualinversion, ruiz2022dreambooth, kumari2022customdiffusion, li2023generate, han2023svdiff, hu2022lora, choi2023custom, kawar2022imagic, voynov2023p+,alaluf2023neural}}, encoder-based~\cite{wei2023elite, jia2023tamingencoder, xu2023prompt, zhou2023enhancing, ma2023subject, ye2023ip}, or hybrid~\extension{\cite{gal2023encoder, chen2023disenbooth, gong2023talecrafter, arar2023domain, li2023blipdiffusion, ruiz2024hyperdreambooth}} approaches.
For example, given several images of a particular \textit{``cat statue''}, 
Textual Inversion~\cite{gal2022textualinversion} learns a new word to describe its appearance - finding a vector in Latent Diffusion Model (LDM)~\cite{rombach2022ldm}'s text embedding space, so that the new word can be composed into new sentences to achieve personalized creation.
Rather than inverting the appearance information (\eg, color, texture), our proposed \RI{} task extracts high-level object \textit{relations} from exemplar images, 
which can be harder as it requires comprehending image compositions and object relationships.

\section{The Relation Inversion Task}

\RI{} aims to extract the common relation \R{} from several exemplar images.
Let $\mathcal{I}\,{=}\, \{I_{1}, I_{2}, ... I_{n}\}$ be a set of exemplar images, and $E_{i,A}$ and $E_{i,B}$ be two dominant entities in image $I_i$. In \RI{}, we assume that the entities in each exemplar image interacts with each other through a common relation $R$. A set of coarse descriptions ${C}~{=}\, \{c_{1}, c_{2}, ... c_{n}\}$ is associated to the exemplar images, where \mbox{``${c}_i\,{=}\,E_{i, A}$ \R{} 
$E_{i, B}$''} denotes the caption corresponding to image $I_i$. Our objective is to optimize the relation prompt \R{} such that the co-existing relation can be accurately represented by the optimized prompt. 

An immediate application of \RI{} is relation-specific text-to-image synthesis. Once the prompt is acquired, one can generate images with novel objects interacting with each other following the specified relation. More generally, this task reveals a new direction of inferring relations from a set of exemplar images. This could potentially inspire future research in representation learning, few-shot learning, visual relation detection, scene graph generation, and many more.

\section{The ReVersion Framework}

\subsection{Preliminaries}

\noindent \textbf{Stable Diffusion.}
Diffusion models are a class of generative models that gradually denoise the Gaussian prior $\bx_T$ to the data $\bx_0$ (\eg, a natural image). 
The commonly used training objective $L_\mathrm{DM}$~\cite{ho2020ddpm} is:
\begin{gather}
    L_\mathrm{DM}(\theta) \defeq \Eb{t, \bx_0, \bepsilon}{ \left\| \bepsilon - \bepsilon_\theta(\bx_t, t) \right\|^2}, \label{eq:training_objective_simple}
\end{gather}
where $\bx_t$ is an noisy image constructed by adding noise $\bepsilon \sim \mathcal{N}(\bzero, \bI)$ to the natural image $\bx_0$, and the network $\bepsilon_\theta(\cdot)$ is trained to predict the added noise.
To sample data $\bx_0$ from a trained diffusion model $\bepsilon_\theta(\cdot)$, we iteratively denoise $\bx_t$ from $t = T$ to $t = 0$ using the predicted noise $\bepsilon_\theta(\bx_t, t)$ at each timestep $t$.

LDM~\cite{rombach2022ldm}, the predecessor of Stable Diffusion, mainly introduced two changes to the vanilla diffusion model~\cite{ho2020ddpm}. First, instead of directly modeling the natural image distribution, LDM models images' projections in autoencoder's compressed latent space. Second, LDM enables text-to-image generation by feeding encoded text input to the UNet~\cite{ronneberger2015unet} $\bepsilon_\theta(\cdot)$. The LDM loss is:
\begin{gather}
    L_\mathrm{LDM}(\theta) \defeq \Eb{t, \bx_0, \bepsilon}{ { \left\| \bepsilon - \bepsilon_\theta(\bx_t, t, \tau_\theta(c)) \right\|^2}}, \label{eq:ldm_loss}
\end{gather}
where $\bx$ is the autoencoder latents for images, and $\tau_\theta(\cdot)$ is the text encoder that encodes the text descriptions $c$ into the text embedding space.
Stable Diffusion extends LDM by training on the larger LAION dataset~\cite{schuhmann2022laion}, with some architectural and training changes.

\noindent \textbf{Inversion on Text-to-Image Diffusion Models.} 
Existing inversion methods focus on appearance inversion. Given several images that all contain a specific entity, they~\cite{gal2022textualinversion, ruiz2022dreambooth, kumari2022customdiffusion} find a text embedding V* for the pre-trained T2I model. The obtained V* can then be used as a new word to generate this entity in different scenarios.

In this work, we aim to capture object relations instead. Given several exemplar images which share a common relation $R$, we aim to find a relation prompt \R{} to capture this relation, such that \mbox{``$E_{A}$ \R{} 
$E_{B}$''} can be used to generate an image where \textit{$E_{A}$} and \textit{$E_{B}$} interact via relation \textit{R}.

\subsection{Relation-Steering Contrastive Learning} 
\label{sec: contrastive}

Recall that our goal is to acquire a relation prompt \R{} that accurately captures the co-existing relation in the exemplar images. 
A basic objective is to reconstruct the exemplar images using \R{}:
\begin{gather}
    \langle{R}\rangle = \argmin_{\langle{r}\rangle} \Eb{t, \bx_0, \bepsilon}{ { \left\| \bepsilon - \bepsilon_\theta(\bx_t, t, \tau_\theta(c)) \right\|^2}}, c~contains~\langle{r}\rangle \label{eq:ti_loss}
\end{gather}
where $\bepsilon \sim \mathcal{N}(\bzero, \bI)$, \R{} is the optimized text embedding, and $\bepsilon_\theta(\cdot)$ is a pre-trained text-to-image diffusion model whose weights are frozen throughout optimization. $\langle{r}\rangle$ is the relation prompt being optimized, and is fed into the pre-trained T2I model as part of the text description $c$. 

However, this pixel-level reconstruction loss mainly focus on reconstructing visual details, without emphasis on object relations. Consequently, we find that directly optimizing with Equation~\ref{eq:ti_loss} could lead the relation prompt $\langle{R}\rangle$ to be more associated with the look of objects  rather than the relation between them, undesirably leaking entity appearance from exemplar images into the generated images, and also causing wrong object relations.

To mitigate this problem, we propose the \textit{``relation-steering contrastive learning''} scheme, leveraging linguistic priors discussed in Section~\ref{sec:intro} 
to emphasis more on object relation during the optimization of $\langle{R}\rangle$. 
Specifically, we sample a set of prepositions as positives and use other Part-of-Speech (POS)' words (\eg, nouns, adjectives) in the text descriptions as negatives to steer the relation prompt towards a relation-dense text embedding subspace, and push it away from appearance-related semantics. Following InfoNCE~\cite{oord2018representation,miech2020end}, we formulate the Steering Loss by:

\begin{gather}
    L_\mathrm{steer} = -log\frac{\sum_{l=1}^{L}{e^{{\langle{r}\rangle}^{\top}\cdot P_{i}^l / \gamma}}}{\sum_{l=1}^{L}{e^{{\langle{r}\rangle}^{\top}\cdot P_{i}^l / \gamma}  + \sum_{m=1}^{M}e^{{\langle{r}\rangle}^{\top}\cdot N_i^m / \gamma}}},
    \label{eq:l_steer}
\end{gather}
where $\langle{r}\rangle$ is the relation embedding, and $\gamma$ is the temperature parameter. $P_i=\{P_i^1, ..., P_i^L\}$ (\ie,~positive samples) refers to a set of a randomly sampled preposition embeddings from basis prepositions (more details provided in Supplementary File) at the $i$-th optimization iteration, and $N_i=\{N_i^1, ..., N_i^M\}$ (\ie,~negative samples) are the embeddings of all other POS' words (\eg, nouns, adjectives) in the exemplars' text descriptions in the current batch. All embeddings are normalized to unit length. We find that our relation-steering contrastive learning scheme can effectively help $\langle{r}\rangle$ to focus on relation and mitigate the appearance leakage problem (see Figure~\ref{fig:ablation_comparison} and Section~\ref{subsec:ablation}).

\subsection{Relation-Focal Importance Sampling}
In the sampling process of diffusion models, high-level semantics usually appear first, and fine details emerge at later stages~\extension{\cite{wang2023diffusion, huang2023collaborative, patashnik2023localizing, liew2022magicmix}}. As our objective is to capture the relation (a high-level concept) in exemplar images, it is undesirable to focus too much on fine-grained visual details (\eg, color, texture) during optimization. Therefore, we further conduct an importance sampling strategy to encourage the learning of high-level relations. Specifically, unlike previous reconstruction objectives, which samples the timestep $t$ from a uniform distribution, we skew the sampling distribution so that a higher probability is assigned to a larger $t$. The Denoising Loss for \textit{``relation-focal importance sampling''} becomes:
\begin{gather}
\begin{split}
    L_\mathrm{denoise} &= \Eb{t\sim f(t),\bx_0, \bepsilon}{ { \left\| \bepsilon - \bepsilon_\theta(\bx_t, t, \tau_\theta(c)) \right\|^2}}, \qquad \\
    f(t)        &= \frac{1}{T}(1 - \alpha \cos{\frac{\pi t}{T}}),
    \label{eq:importance_sampling}
\end{split}
\end{gather}
where $f(t)$ is the importance sampling function, which characterizes the probability density function to sample $t$ from. The skewness of $f(t)$ increases with $\alpha\,{\in}\,(0, 1]$. We set $\alpha=0.5$ throughout our experiments.
The overall optimization objective of the \textbf{\textit{ReVersion Framework}} is:
\begin{gather}
    \langle{R}\rangle = \argmin_{\langle{r}\rangle} (\lambda_\mathrm{steer}L_\mathrm{steer} + \lambda_\mathrm{denoise}L_\mathrm{denoise}), \label{eq:overall_loss}
\end{gather}
where $\lambda_\mathrm{steer}$ and $\lambda_\mathrm{denoise}$ are the weighting factors.

\begin{figure*}[t]
    \centering
     \includegraphics[width=1.00\linewidth]{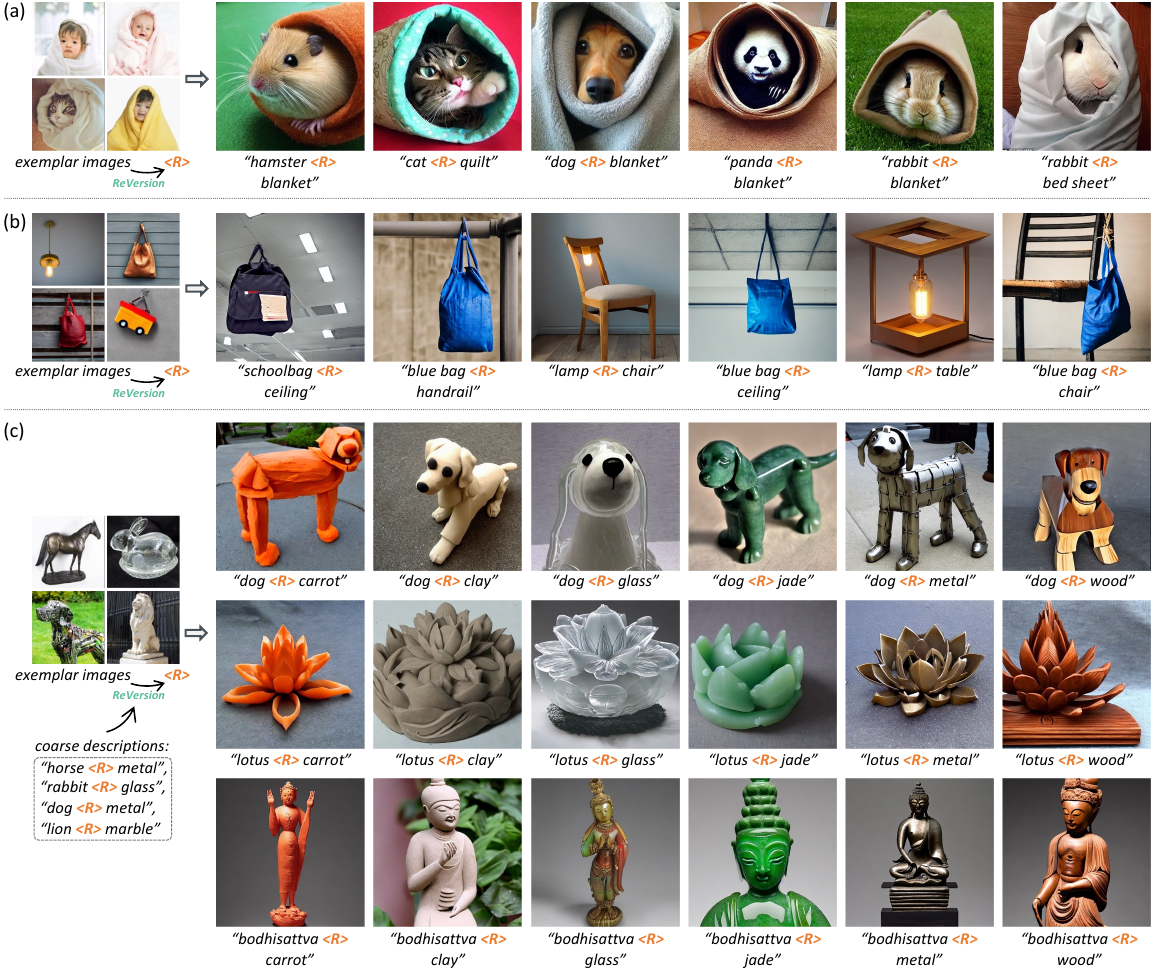}
     \caption{\textbf{Qualitative Results}. Our ReVersion Framework successfully captures the relation that co-exists in the exemplar images, and applies the extracted relation prompt \R{} to compose novel entities. 
     }
     \label{fig:qualitative_results}
\end{figure*}
\section{The ReVersion Benchmark}

To facilitate fair comparison for \textit{Relation Inversion}, we present the \textbf{ReVersion Benchmark}. It consists of diverse \textit{relations} and \textit{entities}, along with a set of well-defined text \textit{descriptions}. This benchmark can be used for conducting qualitative and quantitative evaluations. Additional details are in Supplementary File.

\noindent\textbf{Relations and Entities.} We define ten representative object relations with different abstraction levels, ranging from basic spatial relations (\eg,~\textit{``on top of''}), entity interactions (\eg, \textit{``shakes hands with''}), to abstract concepts (\eg,~\textit{``is carved by''}). A wide range of entities, such as animals, human, household items, are involved to further increase the diversity of the benchmark.

\noindent\textbf{Exemplar Images and Text Descriptions.} For each relation, we collect four to ten exemplar images containing different entities. We further annotate several text templates for each exemplar image to describe them with different levels of details\footnote{For example, a photo of a cat sitting on a box could be annotated as \textbf{1)} \textit{``cat \R{} box"}, \textbf{2)} \textit{``an orange cat \R{} a black box"} and \textbf{3)} \textit{``an orange cat \R{} a black box, with trees in the background"}. Detailed examples will be in the Supplementary File.}. These training templates can be used for the optimization of the relation prompt.

\noindent\textbf{Benchmark Scenarios.} To validate the robustness of the \RI{} methods, we design 100 inference templates composing of different object entities for each of the ten relations. This provides a total of 1,000 inference templates for performance evaluation.

\section{Experiments}

\begin{table}[t]
  \caption{\textbf{Comparisons via Objective Metrics.} We compare our performance against existing methods and ablation variants using objective evaluation metrics.}
  \vspace{-5pt}
  \begin{minipage}{0.50\textwidth} 
      \centering
      \small 
      \subcaption{\textbf{Baseline Comparison.} Performance against several existing methods.}
      \resizebox{0.99\linewidth}{!}{
          \begin{tabular}{>{\hspace{3pt}}l<{\hspace{3pt}}|>{\hspace{3pt}}c<{\hspace{3pt}}|>{\hspace{3pt}}c<{\hspace{3pt}}}
          \Xhline{1pt}
          \textbf{Method} & \textbf{\makecell{Relation Score $\uparrow$}} & \textbf{\makecell{Entity Score $\uparrow$}} \\ \Xhline{1pt}
          Text-to-Image~\cite{rombach2022ldm} &  0.3516 &  0.2896  \\ 
          Textual Inversion~\cite{gal2022textualinversion} &  0.3785 &   0.2679 \\ 
          DreamBooth~\cite{ruiz2022dreambooth} & 0.3576 & \textbf{0.2902} \\ 
          \textbf{Ours} & \textbf{0.3817} & \underline{0.2820} \\
          \Xhline{1pt}
          \end{tabular}
      }
      \label{tab:baseline_quantitative}
  \end{minipage}
  \hspace{0.01\textwidth}
  \begin{minipage}{0.50\textwidth} 
      \centering
      \small
      \vspace{8pt}
      \subcaption{\textbf{Ablation Study}. Steering or importance sampling is removed.}
      \resizebox{0.97\linewidth}{!}{
          \begin{tabular}{>{\hspace{3pt}}l<{\hspace{3pt}}|>{\hspace{3pt}}c<{\hspace{3pt}}|>{\hspace{3pt}}c<{\hspace{3pt}}}
          \Xhline{1pt}
          \textbf{Method} & \textbf{\makecell{Relation Score $\uparrow$}} & \textbf{\makecell{Entity Score $\uparrow$}} \\ \Xhline{1pt}
          Ours w/o Relation-Steering &  0.3748 &  0.2766  \\ 
          Ours w/o Importance Sampling &  0.3464 &   0.2790 \\ 
          \textbf{Ours} & \textbf{0.3817} & \textbf{0.2820} \\
          \Xhline{1pt}
          \end{tabular}
      }
      \label{tab:ablation_quantitative}    
  \end{minipage}
  \hspace{-2pt}
\end{table}

We present qualitative and quantitative results in this section, and more experiments and analysis are in the Supplementary File.
We adopt Stable Diffusion~\cite{rombach2022ldm} for all experiments since it achieves a good balance between quality and speed. We generate images at $512\times512$ resolution.

\subsection{Comparison Methods}

\noindent \textbf{Text-to-Image Generation using Stable Diffusion~\cite{rombach2022ldm}}. We use the original Stable Diffusion 1.5 as the text-to-image generation baseline. Since there is no ground-truth textual description for the relation in each set of exemplar images, we use natural language that can best describe the relation to replace the \R{} token.
For example, in Figure~\ref{fig:baseline_comparison} (a), the co-existing relation in the reference images can be roughly described as \textit{``is painted on"}. Thus, we use it to replace the \R{} token in the inference template \textit{``Spiderman \R{} building''}, resulting in a sentence \textit{``Spiderman is painted on building''}, which is then used as the text prompt for generation.

\noindent \textbf{Textual Inversion~\cite{gal2022textualinversion}}.
For fair comparison with our method developed on Stable Diffusion 1.5, we use the \textit{diffusers}~\cite{diffusers} implementation of Textual Inversion~\cite{gal2022textualinversion} on Stable Diffusion 1.5. Based on the default hyper-parameter settings, we tuned the learning rate and batch size for its optimal performance on our \RI{} task. We use Textual Inversion's LDM objective to optimize \R{} for 3000 iterations, and generate images using the obtained \R{}.

\noindent \textbf{DreamBooth~\cite{ruiz2022dreambooth}}.
We use \textit{diffusers} implementation of DreamBooth on Stable Diffusion 1.5.
To adapt DreamBooth to our \RI{} task for fair comparison, we made three modifications to the original implementation. First, instead of using the original training template like ``A photo of V* dog'', we explicitly inject the word ``relation'' into the text template to help DreamBooth focus on relation instead of entity, thereby using ``A photo of \R{} \textit{relation}'' to fine-tune the model. Second, the class-specific prior preservation loss is implemented with a text prompt ``A photo of \textit{relation}'' to avoid overfitting or language drift. Third, to align with fine-tuning stage's template, the template ``Entity A is \textit{in \R{} relation} with Entity B'' is used during inference. 

\subsection{Qualitative Comparisons}

\noindent\textbf{Our Results.}
In Figure~\ref{fig:qualitative_results}, we provide the generation results using \R{} inverted by ReVersion. We observe that our framework is capable of 1) synthesizing the entities in the inference template and 2) ensuring that entities follow the relation co-existing in the exemplar images. We provide additional qualitative results in the Supplementary File due to space constraint.

\begin{figure*}[t]
  \centering
  \includegraphics[width=0.93\linewidth]{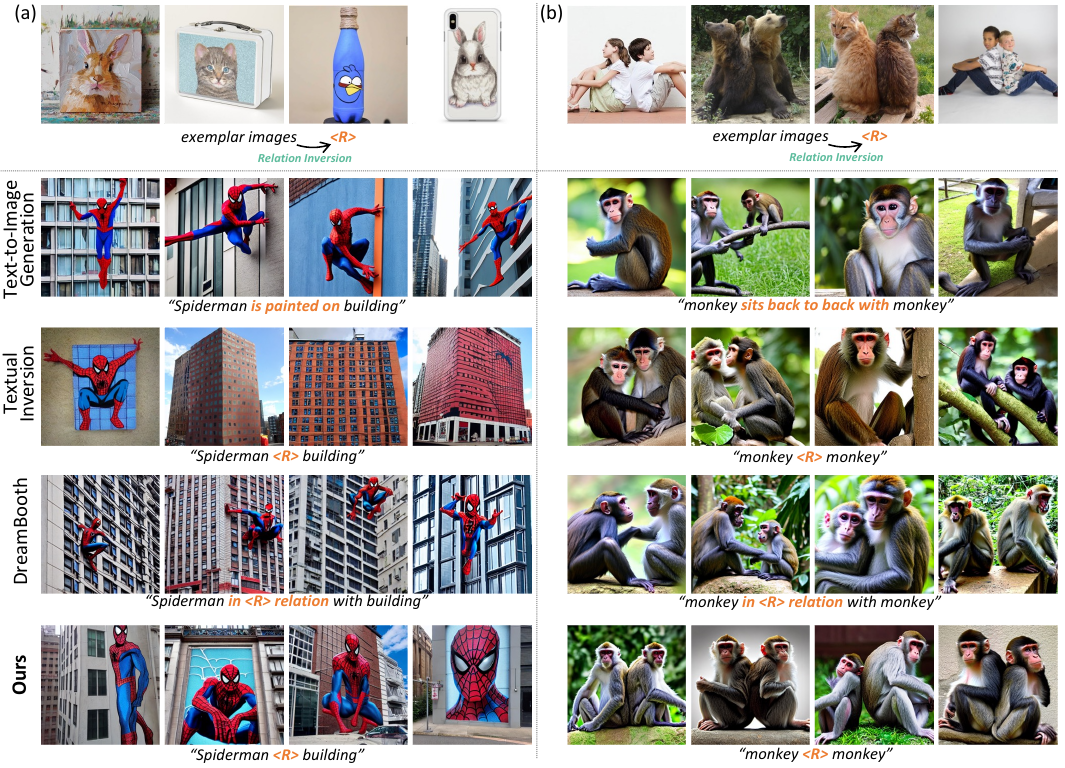}
  \vspace{-5pt}
  \caption{
        \textbf{Qualitative Comparisons with Existing Methods}. 
        Our method can generate entity and relation accurately. ``Text-to-Image Generation'' and ``DreamBooth'' can correctly generate entities described in text prompt, but fail to compose them following the desired relation. ``Textual Inversion'' suffers from appearance leakage (\eg, \R{} unexpectedly capturing the canvas in exemplar images), thus resulting in low entity accuracy (\eg, cannot generate spiderman and building simultaneously).
   }
   \label{fig:baseline_comparison}
  \end{figure*}

\noindent\textbf{Comparison of Relation Accuracy.}
Figure~\ref{fig:baseline_comparison} shows qualitative comparisons with existing methods. We compare our method with 1) Text-to-Image Generation via Stable Diffusion~\cite{rombach2022ldm}, 2) Textual Inversion~\cite{gal2022textualinversion}, and 3) DreamBooth~\cite{ruiz2022dreambooth}.
In Figure~\ref{fig:baseline_comparison} (a), although ``Text-to-Image Generation'' and ``DreamBooth'' successfully generate both entities (Spiderman and building), they fail to \textit{paint} Spiderman on the building as the exemplar images do. They severely rely on the bias between two entities: Spiderman usually \textit{climbs}/\textit{jumps} on the buildings, instead of being \textit{painted} onto the buildings. Similarly, in Figure~\ref{fig:baseline_comparison} (b), although all methods in comparison can generate at least one monkey, the relation between generated monkeys does not follow the \textit{``back to back''} relation in the exemplar images. In contrast,
Our ReVersion Framework does not have this problem. 

\begin{figure*}[t]
  \centering
  \includegraphics[width=0.90\linewidth]{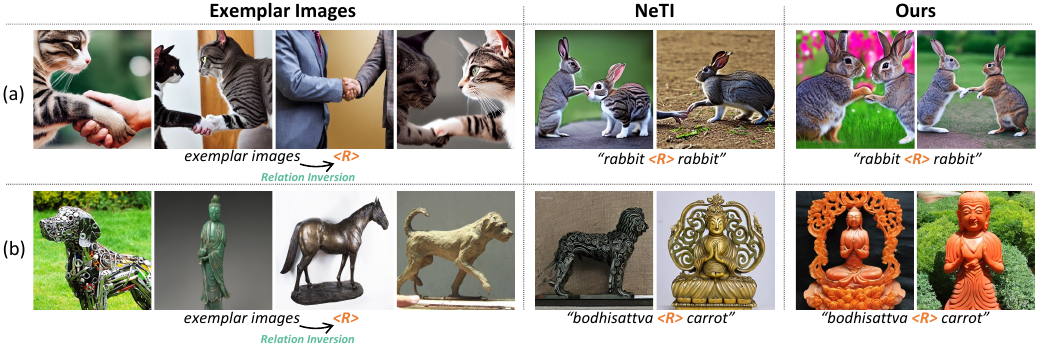}
  \vspace{-10pt}
   \caption{
      \extension{
        \textbf{Comparisons with Newer Method}. NeTI~\cite{alaluf2023neural} demonstrates \textit{some degree of effectiveness} for relation inversion, attributed to its adaptive adjustment at different network layers and denoising timesteps. For example, in (a) where \R{} denotes ``shaking hands'', NeTI successfully rendered rabbits extending their hands, trying to engage in the ``shake hands'' behaviour. However, NeTI is still prone to \textit{texture leakage}. For instance: (a) The striped patterns of cat fur from the exemplar images are unintentionally transferred to the rabbit fur in NeTI's outputs. (b) With the ``carved by'' relation, the metal dog appearance in the exemplar images is unintentionally captured by NeTI, resulting in images resembling a metal animal even when the text prompt is ``bodhisattva \R{} carrot''. Our relation steering is essential to help \R{} focus on the relation rather than the appearance, thereby producing results without texture leakage.
      }
   }
   \label{fig:fig_neti}
   \vspace{-35pt}
\end{figure*}

\noindent\textbf{Entity Leakage in Existing Methods.}
In Textual Inversion, entities in the exemplar images like canvas are leaked to \R{}, such that the generated image shows a Spiderman on the canvas even when the word \textit{``canvas''} is not in the inference prompt (see Figure~\ref{fig:baseline_comparison} (a)). In DreamBooth, the ``basket'' in exemplar images sometimes leak to the generated images (see Figure~\ref{fig:fig_dreambooth_leak}).
\extension{
  In Figure~\ref{fig:fig_neti}, we include comparisons with NeTI~\cite{alaluf2023neural} and also discuss its entity leakage problem.
}

\subsection{Quantitative Comparisons via Human Evaluation}

\begin{table}[t]
  \centering
  \caption{\textbf{Comparison with Existing Methods (Human Preference).} Percentage of votes where users favor our results vs. comparison methods. Our method outperforms the baselines under all metrics.}
    \small 
    \resizebox{0.99\linewidth}{!}{
        \begin{tabular}{>{\hspace{3pt}}l<{\hspace{3pt}}|>{\hspace{3pt}}c<{\hspace{3pt}}|>{\hspace{3pt}}c<{\hspace{3pt}}|>{\hspace{3pt}}c<{\hspace{3pt}}}
        \Xhline{1pt}
        \textbf{Method} & \textbf{Relation Accuracy} & \textbf{Entity Accuracy} & \textbf{Overall Quality} \\ \Xhline{1pt}
        Text-to-Image Generation~\cite{rombach2022ldm} & 6.45\% & 10.32\% & 9.68\%  \\ 
        Textual Inversion~\cite{gal2022textualinversion} & 6.13\% & 5.81\% & 5.16\% \\ 
        DreamBooth~\cite{ruiz2022dreambooth} & 18.39\% & 18.39\% & 19.03\% \\ 
        \textbf{Ours} & \textbf{69.03\%} & \textbf{65.48\%} & \textbf{66.13\%}  \\
        \Xhline{1pt}
      \end{tabular}
    }
  \label{tab:quantitative_human}
\end{table}
\begin{table}[t]
  \centering
  \vspace{5pt}
  \caption{\textbf{Ablation Study (Human Preference)}. Suppressing relation-steering or importance sampling introduces performance drops, which shows the necessity of both relation-steering and importance sampling.
  }
    \small %
    \resizebox{0.99\linewidth}{!}{    
        \begin{tabular}{>{\hspace{3pt}}l<{\hspace{20pt}}|>{\hspace{3pt}}c<{\hspace{3pt}}|>{\hspace{3pt}}c<{\hspace{3pt}}|>{\hspace{3pt}}c<{\hspace{3pt}}}
        \Xhline{1pt}
        \textbf{Method} & \textbf{Relation Accuracy} & \textbf{Entity Accuracy} & \textbf{Overall Quality} \\ \Xhline{1pt}
        w/o Relation-Steering & 11.20\% & 10.90\% & 13.31\%  \\ 
        w/o Importance Sampling & 11.20\% & 13.62\% & 7.14\% \\ 
        \textbf{Ours} & \textbf{77.60\%} & \textbf{75.48\%} & \textbf{79.55\%}  \\
        \Xhline{1pt}
        \end{tabular}
    }
  \label{tab:ablation_human}
\end{table}

We conduct user studies with 68 human evaluators to assess the performance of our ReVersion Framework on the \RI{} task. We sampled 20 groups of images. Each group has images generated by different methods or ablation variants. For each group, apart from the generated images, the following information is presented: 1) exemplar images of a particular relation, 2) text description of the exemplar images. We then ask the evaluators to vote for the best generated image with respect to the following metrics.

\begin{figure*}[t]
  \centering
  \vspace{10pt}
   \includegraphics[width=1.00\linewidth]{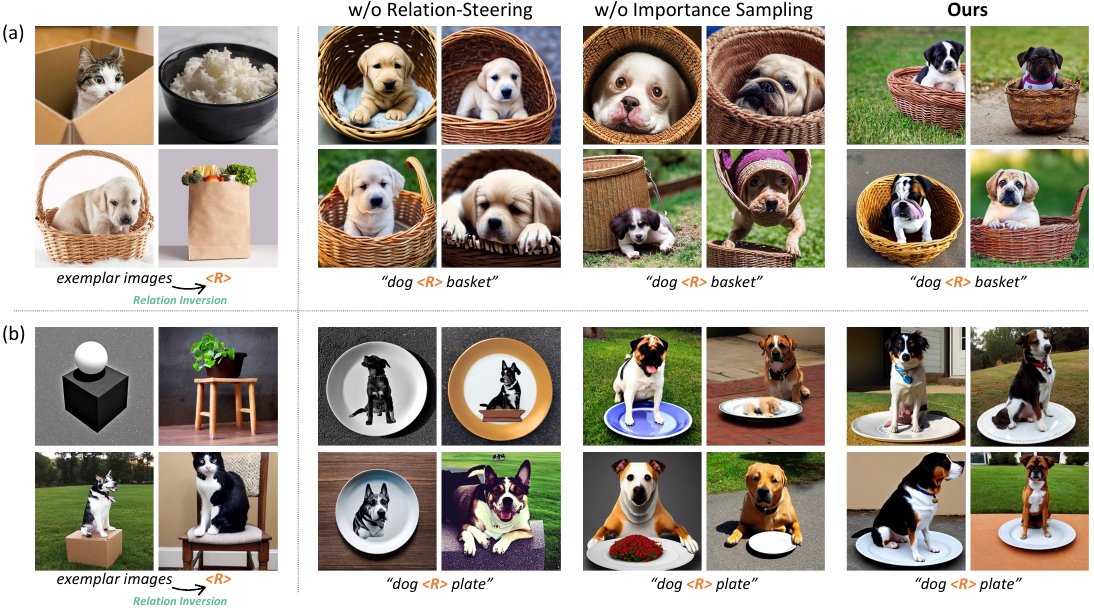}
   \caption{
        \textbf{Ablation Study (Qualitative)}. Without relation-steering, \R{} suffers from appearance leak (\eg, white puppy in (a), gray background in (b)) and inaccurate relation capture (\eg, dog not being on top of plate in (b)). Without importance sampling, \R{} focuses on lower-level visual details (\eg, rattan around puppy in (a)) and misses high-level relations.
   }
   \label{fig:ablation_comparison}
   \vspace{5pt}
\end{figure*}

\noindent \textbf{Relation Accuracy}.
Human evaluators are asked to evaluate whether the relations of the two entities in the generated image are consistent with the relation co-existing in the exemplar images.

\noindent \textbf{Entity Accuracy}.
Given an inference template in the form of \textit{``$\langle$Entity A$\rangle$ \R{} $\langle$Entity B$\rangle$''}, we ask evaluators to determine whether \textit{$\langle$Entity A$\rangle$} and \textit{$\langle$Entity B$\rangle$} are both authentically generated in each image.

\noindent \textbf{Overall Quality}.
Human evaluators are asked to assess the overall performance on the ReVersion task, considering both the alignment of relation and entity, and the image quality.

\noindent
Table \ref{tab:quantitative_human} shows our method clearly obtains better results under all three metrics.

\subsection{Quantitative Comparisons via Objective Metrics}

\begin{figure}[t]
  \centering
  \vspace{10pt}
   \includegraphics[width=1.00\linewidth]{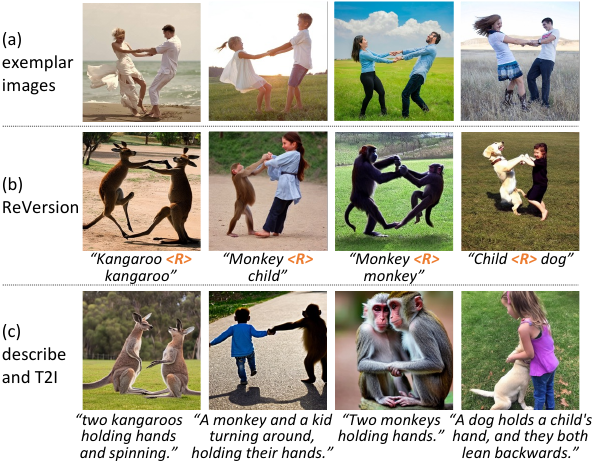}
   \caption{
    \textbf{ReVersion for Complicated Relation.}
     \textbf{\textit{(a) Exemplar images.}} In each exemplar image, people exhibit the similar relation of ``holding hands, leaning backwards''.
     \textbf{\textit{(b) Ours.}} ReVersion effectively captures this relation by \R{} and successfully applies it to new entities.
     \textbf{\textit{(c) Describe and T2I.}} The ``first describe the relation, then use text-to-image'' approach struggles to accurately represent such complex relation in newly synthesized images.
 }
   \label{fig:fig_paper_complicated_relation}
\end{figure}
\begin{figure}[t]
  \centering
  \vspace{10pt}
   \includegraphics[width=1.00\linewidth]{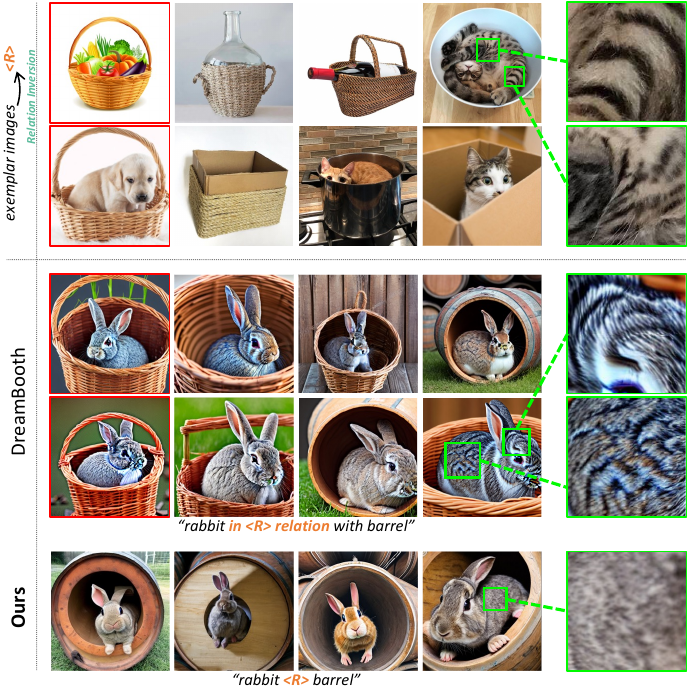}
   \caption{
    \textbf{Appearance Leakage of DreamBooth}. 
    \textbf{\textit{(a) Entity Leakage (Red Boxes):}} 
    The basket from the exemplar images significantly leaks into images generated by DreamBooth. In contrast, our approach avoids this issue of entity leakage.
    \textbf{\textit{(b) Texture Leakage (Green Boxes):}} 
    While DreamBooth accurately generates the entity ``rabbit'', it encounters texture leakage from the exemplar images. That is, stripe patterns of cat fur texture (marked with green boxes) unintentionally transfer to the rabbit's fur in DreamBooth's outputs. Our method, in contrast, is free from such texture leakage.
}
   \label{fig:fig_dreambooth_leak}
\end{figure}

We devise automatic metrics to objectively evaluate ``relation accuracy'' and ``entity accuracy'', which are briefly introduced below. More implementation details of the objective metrics will be detailed in the Supplementary File.
For comparison experiments, we use the 1,000 inference templates in the ReVersion Benchmark for all relations, and generate 10 images using each template. 

\noindent\textbf{Relation Score.} 
We use PSGFormer~\cite{yang2022psg}, a pre-trained scene-graph generation network, to extract the relation features for relation accuracy evaluation.
Table~\ref{tab:baseline_quantitative} shows that our method outperforms all existing methods in comparison.

\noindent\textbf{Entity Score.} 
We use CLIP~\cite{radford2021clip} score to calculate the alignment between the entity types in the text prompt versus the generated entities.
Table~\ref{tab:baseline_quantitative} shows that our method  outperforms Textual Inversion in terms of entity accuracy. This is because the \R{} learned by Textual Inversion contains leaked entity information, which distracts the model from generating the desired ``$E_{A}$'' and ``$E_{B}$''. Our steering loss effectively prevents entity information from leaking into \R{}, allowing for accurate entity synthesis. Furthermore, our approach achieves comparable entity score with ``Text-to-Image Generation'' and ``DreamBooth'', and significantly surpasses them in terms of relation score. 
It is worth mentioning that the CLIP-based metrics mainly focus on whether the correct class of object is generated, and does not fully take the pixel-level object quality into account. For example, as shown in Figure~\ref{fig:fig_dreambooth_leak}, the stripe textures of cat fur in exemplar images often leak to \R{}, resulting in unrealistic textures in generated rabbits.

\subsection{Ablation Study}
\label{subsec:ablation}

From both Table~\ref{tab:ablation_human} (human evaluation) and Table~\ref{tab:ablation_quantitative} (objective metrics), we observe that removing steering or importance sampling results in deterioration in both relation accuracy and entity accuracy. This corroborates our observations that 1) relation-steering effectively guides \R{} towards the relation-dense regions and disentangles \R{} away from exemplar entities, and 2) importance sampling emphasizes high-level relations over low-level details, aiding \R{} to be relation-focal. We further show qualitatively the necessity of both modules in Figure~\ref{fig:ablation_comparison}.

\noindent \textbf{Effectiveness of Relation-Steering.}
In ``w/o Relation-Steering'', we remove the Steering Loss $L_\mathrm{steer}$ in the optimization process. As shown in Figure~\ref{fig:ablation_comparison}~(a), the appearance of the white puppy in the lower-left exemplar image is leaked into \R{}, resulting in similar puppies in the generated images. In Figure~\ref{fig:ablation_comparison}~(b), many appearance elements are leaked into \R{}, such as the gray background, the black cube, and the husky dog. The dog and the plate also do not follow the relation of \textit{``being on top of''} as shown in exemplar images. Consequently, the images generated via \R{} do not present the correct relation and introduced unwanted leaked imageries.

\noindent \textbf{Effectiveness of Importance Sampling.}
We replace our relation-focal importance sampling with uniform sampling, and observe that \R{} pays too much attention to low-level details rather than high-level relations. For instance, in Figure~\ref{fig:ablation_comparison}~(a) ``w/o Importance Sampling'', the basket rattan wraps around puppy's head in the same way as the exemplar image, instead of containing the puppy inside.

\subsection{Further Analysis}

\begin{figure*}[t]
  \centering
      \includegraphics[width=0.99\textwidth]{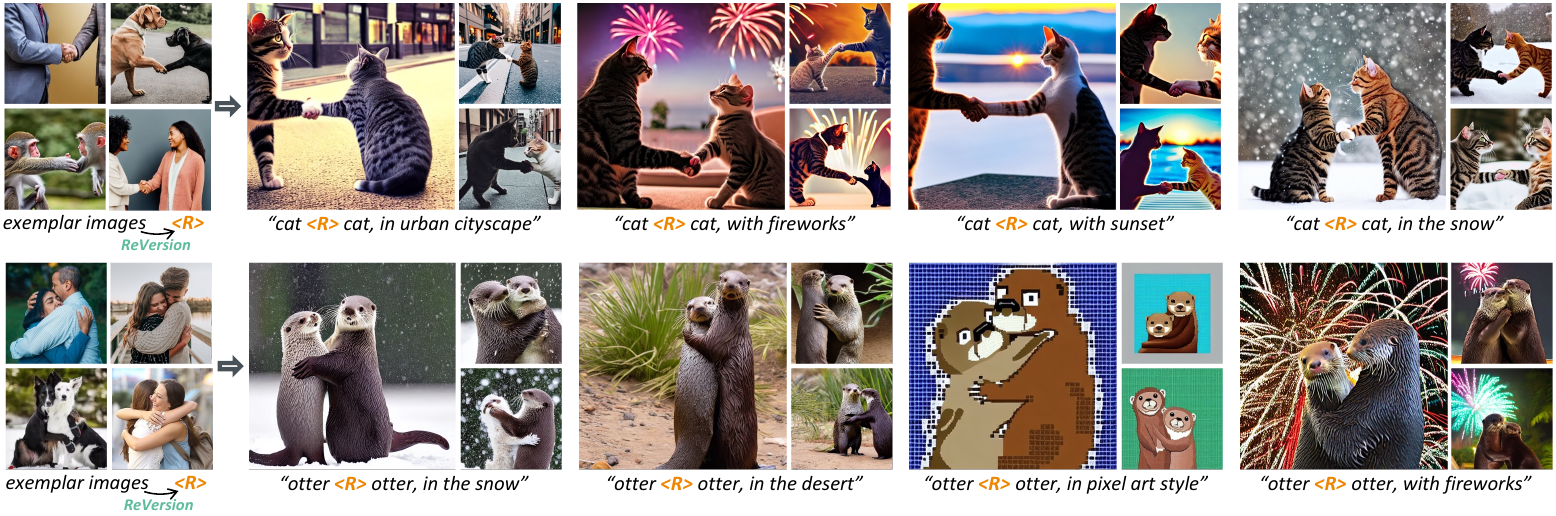}
      \caption{
          \textbf{ReVersion for Diverse Styles and Backgrounds}. The \R{} inverted by ReVersion can be applied robustly to relate entities under diverse backgrounds or styles.
      }
      \label{fig:fig_paper_styles}
  \end{figure*}

\noindent \textbf{Diverse Styles and Backgrounds}. As shown in Figure~\ref{fig:fig_paper_styles}, the \R{} inverted by ReVersion can be applied robustly to relate entities in scenes with diverse backgrounds or styles.

\noindent \textbf{More Comparisons on Complicated Relation}.
Some relations are hard to accurately express by text, 
or the description of such relation may be complex and difficult for the text-to-image generation model to effectively comprehend. For the relation shown in Figure~\ref{fig:fig_paper_complicated_relation} (a), our method (Figure~\ref{fig:fig_paper_complicated_relation} (b)) effectively captures these relations using \R{} and applies them to new entities. In Figure~\ref{fig:fig_paper_complicated_relation} (c), we engage four human subjects to observe the exemplar images in (a) and describe scenes where these relations are applied to new entities (detailed process in Supplementary File). Subsequently, we utilize text-to-image (T2I) to synthesize images based on these human descriptions. The results demonstrate that this ``describe and T2I'' approach struggles to accurately represent such complex relations in the newly synthesized images.

\subsection{Limitations and Potential Societal Impacts}

\noindent \textbf{Limitations}.
Our performance is dependent on the generative capabilities of Stable Diffusion. For instance, it might produce sub-optimal synthesis results for entities that Stable Diffusion struggles at, such as human body and human face. We discuss limitations of ``human synthesis'' and ``concept blending'' in detail in the Supplementary File with qualitative examples.

\noindent\textbf{Potential Negative Societal Impacts}.
The entity relational composition capabilities of \textit{ReVersion} could be applied maliciously on real human figures. Additional potential impacts are discussed in the Supplmentary File in depth.

\section{Conclusion}

In this work, we take the first step forward and propose the \textbf{\RI{}} task, which aims to learn a relation prompt to capture the relation that co-exists in multiple exemplar images. In our \textbf{\textit{ReVersion Framework}}, we use \textit{relation-steering contrastive learning} scheme to effectively guide the relation prompt towards relation-dense regions in the text embedding space, and our \textit{relation-focal importance sampling} scheme shift the focus from visual details to high-level relations. We also contribute the \textbf{\textit{ReVersion Benchmark}} for performance evaluation. Our proposed \textbf{\RI{}} task would be a good inspiration for future works in various domains such as generative model inversion, representation learning, few-shot learning, visual relation detection, and scene graph generation.
\balance

\vspace{20pt}
\begin{acks}
    This study is supported by the Ministry of Education, Singapore, under its MOE AcRF Tier 2 (MOET2EP20221- 0012), NTU NAP, and under the RIE2020 Industry Alignment Fund – Industry Collaboration Projects (IAF-ICP) Funding Initiative, as well as cash and in-kind contribution from the industry partner(s).
\end{acks}

\clearpage
\bibliographystyle{ACM-Reference-Format}
\bibliography{main}


\clearpage
\appendix
\section*{Supplementary}
\renewcommand\thesection{\Alph{section}}
\renewcommand\thefigure{A\arabic{figure}}
\renewcommand\thetable{A\arabic{table}}
\setlength{\parskip}{3pt}

\noindent


In this \textbf{\textit{supplementary file}}, we provide more experimental details in Section~\ref{sec:experiment}, and elaborate on the ReVersion Benchmark details in Section~\ref{sec:benchmark_details}. We then provide further explanations on basis prepositions in Section~\ref{sec:prepositions}. We also discuss our limitations in Section~\ref{sec:limitations}, and the potential societal impacts of our work in Section~\ref{sec:impact}.  At the end of the supplementary file, we show various qualitative results of ReVersion in Section~\ref{sec:qualitative}.

\section{More Experimental Details}
\label{sec:experiment}

\begin{figure*}[h!t]
    \centering
    \vspace{10pt}
    \includegraphics[width=0.60\linewidth]{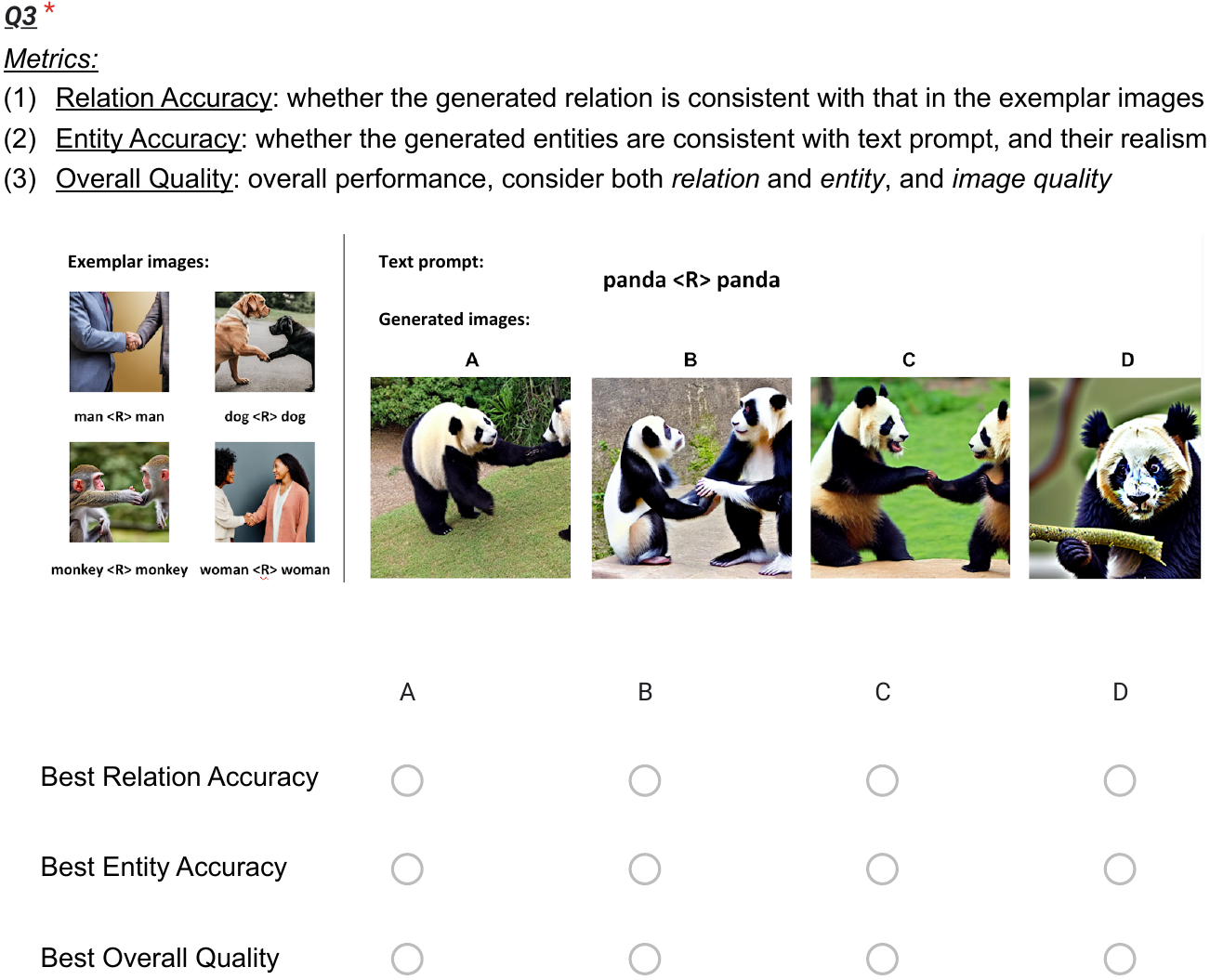}
     \caption{
          \textbf{Example of Human Evaluation}. This is a screenshot of a user study question distributed to human evaluators. The order of different methods (\ie, $A$, $B$, $C$, and $D$) is randomized. Human evaluators are provided with the exemplar images, text prompt, and generated images. They are asked to vote for the best generated image among $A$, $B$, $C$, and $D$, for the three metrics (\ie, \textit{Relation Accuracy / Entity Accuracy / Overall Quality}) respectively.
     }
     \vspace{30pt}
     \label{fig_supp_user_study}
  \end{figure*}

  \begin{figure*}[h!t]
    \centering
     \includegraphics[width=0.60\linewidth]{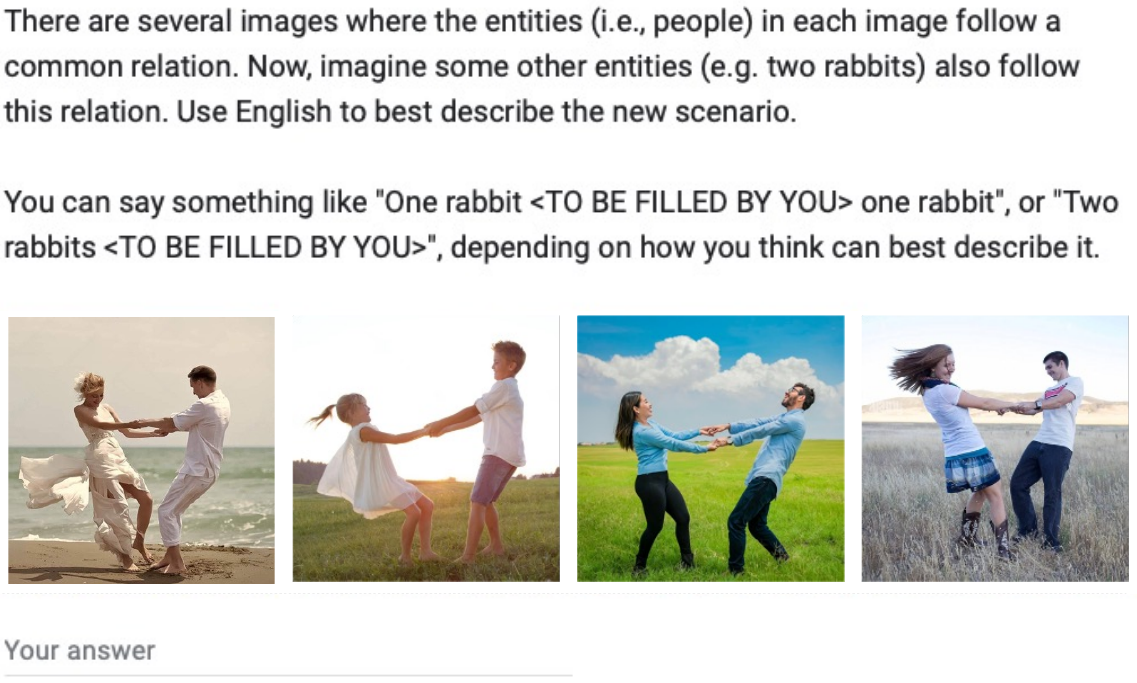}
     \caption{
          \textbf{Human Description of Relation}. This is a screenshot of a user study question distributed to human subjects. The human subjects are asked to observe the exemplar images and identify the co-existing relation in the exemplar images. They are then asked to use natural language to describe the relation. The description will then be used for the ``Describe and T2I'' baseline.
     }
     \vspace{10pt}
     \label{fig_supp_difficult_relation_user_study}
     \vspace{10pt}
  \end{figure*}

In this section, we provide more experimental details.

\subsection{Implementation Details of ReVersion}
\label{sec:implementation_details}

We introduce the implementation details of the ReVersion Framework.
Our framework is built on top of the \textit{diffusers}~\cite{diffusers} implementation of Stable Diffusion~\cite{rombach2022ldm} 1.5.
All experiments are conducted on $512{\times}512$ image resolution.
In Equation 4, the temperature parameter $\gamma$ in the steering loss $L_\mathrm{steer}$ is set as 0.07, following~\cite{he2020momentum}. In each iteration, 8 positive samples are randomly selected from the basis preposition set (see Table~\ref{tab:preposition_words}). 
In Equation 6, to ensure that the numerical values $\lambda_\mathrm{denoise}L_\mathrm{denoise}$ and $\lambda_\mathrm{steer}L_\mathrm{steer}$ are in comparable order of magnitude, we set $\lambda_\mathrm{denoise}=1.0$ and  $\lambda_\mathrm{steer}=0.01$. 
During the optimization process, we first initialize our relation prompt \R{} using the word \textit{``and''}, then optimize the prompt using the AdamW~\cite{loshchilov2017adamw} optimizer for 3,000 steps, with learning rate $2.5{\times}10^{-4}$ and batch size 2. 
During the inference process, we use classifier-free guidance for all experiments including the baselines and ablation variants, with a constant guidance weight 7.5.

\subsection{Human Evaluation}

We introduce the implementation details of the user studies in the main paper's Section 6.3.

Figure~\ref{fig_supp_user_study} is a screenshot of the user study form we distributed for main paper's Table 1, namely ``Comparing with Existing Methods''. We employ preference 
voting to differentiate the performance of different methods.
To ensure unbiased responses, the order of different methods' results is randomized. That is, the orders of generated images $A$, $B$, $C$, and $D$ are random and different in each question. 
For main paper's Table 1, ``Comparison with Existing Methods'', four methods are in comparison, so there are four choices: $A$, $B$, $C$, and $D$.
For main paper's Table 2, ``Ablation Study'', three methods are in comparison, so there are three choices: $A$, $B$, and $C$.

\subsection{Objective Evaluation Metrics}

We introduce the implementation details of the objective metrics used in the main paper's Section 6.4.

\noindent\textbf{Relation Score.}
We devise an objective evaluation metric to measure the quality and accuracy of the inverted relation. To do this, we train relation classifiers that categorize the ten relations in our ReVersion benchmark. We then use these classifiers to determine whether the entities in the generated images follow the specified relation. We employ PSGFormer~\cite{yang2022psg}, a pre-trained scene-graph generation network, to extract the relation feature vectors from a given image. The feature vectors are averaged-pooled and fed into linear SVMs for classification.
We calculate the \textit{Relation Score} as the percentage of generated images that follow the relation class in the exemplar images. 

\noindent\textbf{Entity Score.}
To evaluate whether the generated image contains the entities specified by the text prompt, we compute the CLIP score~\cite{radford2021clip} between a revised text prompt and the generated image, which we refer to as the \textit{Entity Score}.
CLIP~\cite{radford2021clip} is a vision-language model that has been trained on large-scale datasets. It uses an image encoder and a text encoder to project images and text into a common feature space. The CLIP score is calculated as the cosine similarity between the normalized image and text embeddings. A higher score usually indicates greater consistency between the output image and the text prompt.
In our approach, we calculate the CLIP score between the generated image and the revised text prompt ``$E_{A}$\textit{,} $E_{B}$'', which only includes the entity information. 

\subsection{Implementation of ``Describe and Text-to-Image''}
In main paper's Section 6.6 and Figure 6, we compared our method against the ``Describe and Text-to-Image (T2I)'' approach. We provide detailed process in Figure~\ref{fig_supp_difficult_relation_user_study}.

\section{ReVersion Benchmark Details}
\label{sec:benchmark_details}

In this section, we provide the details of our ReVersion Benchmark. The full benchmark will be publicly available.

\begin{figure*}[h!t]
  \centering
   \includegraphics[width=0.99\linewidth]{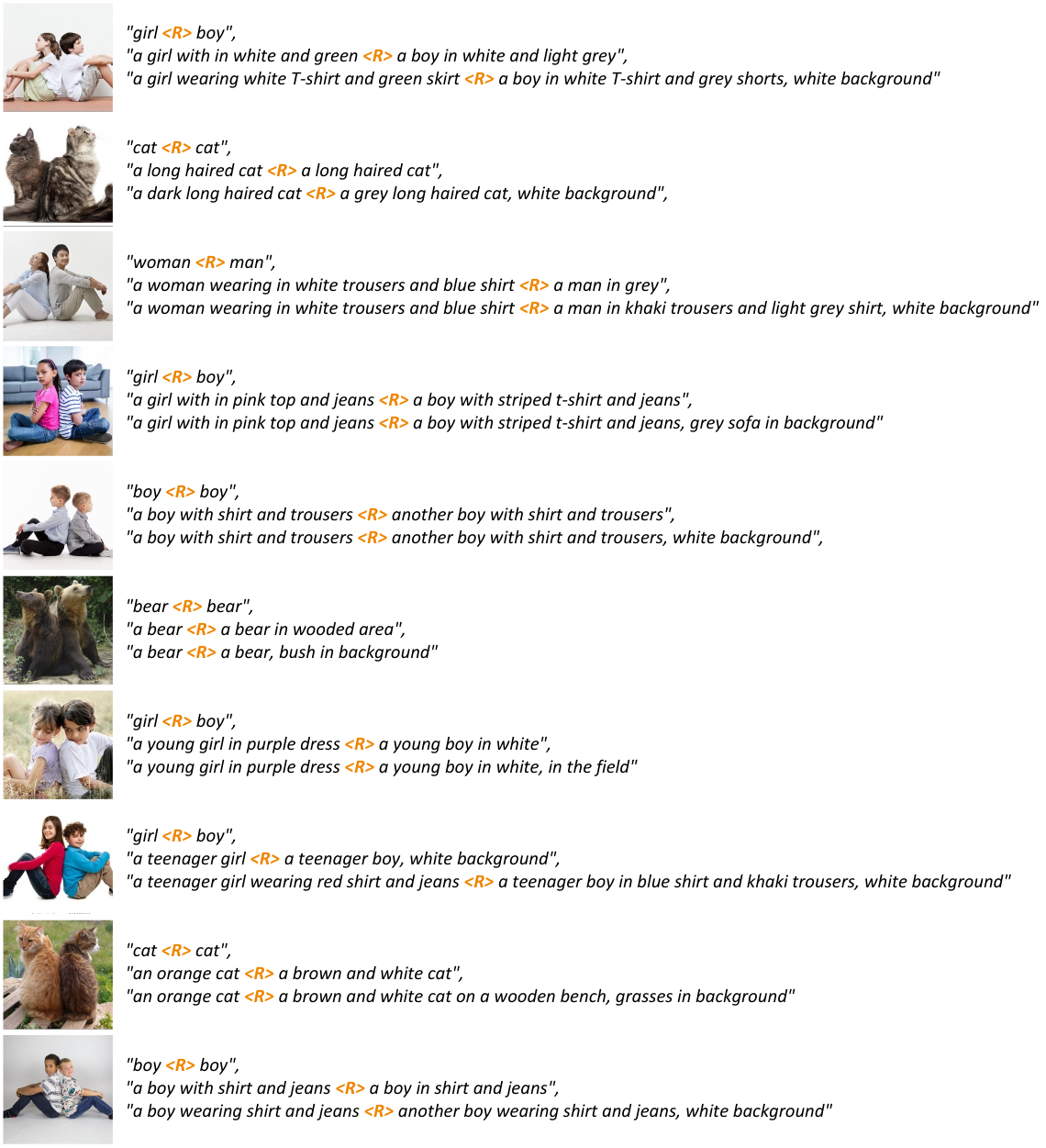}
   \caption{\textbf{Benchmark Sample}. We present \textit{exemplar images} and \textit{text descriptions} that illustrate the relation where ``$E_{A}$ \textit{sits back to back with} $E_{B}$''. The \textit{exemplar images} feature both human figures and animals to demonstrate the invariant \textit{``back to back''} relationship in various scenarios. The \textit{text descriptions} are provided at several levels, ranging from simple class name mentions to detailed descriptions of the entities and their surroundings. During optimization, the \R{} in each description will be replaced with the learnable relation prompt.
   }
   \vspace{10pt}
   \label{fig_supp_benchmark}
\end{figure*}

\subsection{Relations}
To benchmark the Relation Inversion task, we define ten diverse and representative object relations as follows:

\begin{itemize}
    \setlength\itemsep{0em}
    \item $E_{A}$ \textbf{\textit{is painted on (the surface of)}} $E_{B}$
    \item $E_{A}$ \textbf{\textit{is carved by / is made of the material of}} $E_{B}$
    \item $E_{A}$ \textbf{\textit{shakes hands with}} $E_{B}$
    \item $E_{A}$ \textbf{\textit{hugs}} $E_{B}$
    \item $E_{A}$ \textbf{\textit{sits back to back with}} $E_{B}$
    \item $E_{A}$ \textbf{\textit{is contained inside}} $E_{B}$
    \item $E_{A}$ \textbf{\textit{on / is on top of}} $E_{B}$
    \item $E_{A}$ \textbf{\textit{is hanging from}} $E_{B}$
    \item $E_{A}$ \textbf{\textit{is wrapped in}} $E_{B}$
    \item $E_{A}$ \textbf{\textit{rides (on)}} $E_{B}$
\end{itemize}
where $E_{A}$ and $E_{B}$ are the two entities that follow the specified relation. It is worth mentioning that the relations can be best described by the exemplar images, and the text descriptions provided above are simply approximated summarizations of the true relations.

\subsection{Exemplar Images}
A wide range of entities, such as animals, human, household items, are involved to further increase the diversity of the benchmark. In Figure~\ref{fig_supp_benchmark}, we show the \textit{exemplar images} and \textit{text descriptions} for the relation ``$E_{A}$ \textit{sits back to back with} $E_{B}$''. The exemplar images contain both human figures and animals to emphasize the invariant \textit{``back to back''} relation in different scenarios.

\subsection{Text Descriptions}
As shown in Figure~\ref{fig_supp_benchmark}, the \textit{text descriptions} for each image contains several levels, from short sentences which only mention the class names, to complex and comprehensive sentences that describe each entity and the scene backgrounds. The \R{} in each description will be replaced by the learnable relation prompt during optimization.

\subsection{Inference Templates}
To evaluate the performance of relation inversion methods, we devise 100 inference templates for each relation. The inference templates contains diverse entity combinations to test the robustness and generalizability of the inverted relation \R. To quantitatively evaluate relation inversion performance, we use each inference template to synthesize 10 images, resulting in a total of 1,000 synthesized images for each inverted \R.

Below, we show the 100 inference templates for the relation ``$E_{A}$ \textit{sits back to back with} $E_{B}$'':
\vspace{-0.4em}
\begin{itemize}
    \setlength\itemsep{0em}
    \item \textit{man \R{} man, man \R{} woman, man \R{} child, man \R{} cat, man \R{} rabbit, man \R{} monkey, man \R{} dog, man \R{} hamster, man \R{} kangaroo, man \R{} panda, }
    \item \textit{woman \R{} man, woman \R{} woman, woman \R{} child, woman \R{} cat, woman \R{} rabbit, woman \R{} monkey, woman \R{} dog, woman \R{} hamster, woman \R{} kangaroo, woman \R{} panda, } 
    \item \textit{child \R{} man, child \R{} woman, child \R{} child, child \R{} cat, child \R{} rabbit, child \R{} monkey, child \R{} dog, child \R{} hamster, child \R{} kangaroo, child \R{} panda,  }
    \item \textit{cat \R{} man, cat \R{} woman, cat \R{} child, cat \R{} cat, cat \R{} rabbit, cat \R{} monkey, cat \R{} dog, cat \R{} hamster, cat \R{} kangaroo, cat \R{} panda,  }
    \item \textit{rabbit \R{} man, rabbit \R{} woman, rabbit \R{} child, rabbit \R{} cat, rabbit \R{} rabbit, rabbit \R{} monkey, rabbit \R{} dog, rabbit \R{} hamster, rabbit \R{} kangaroo, rabbit \R{} panda,  }
    \item \textit{monkey \R{} man, monkey \R{} woman, monkey \R{} child, monkey \R{} cat, monkey \R{} rabbit, monkey \R{} monkey, monkey \R{} dog, monkey \R{} hamster, monkey \R{} kangaroo, monkey \R{} panda,  }
    \item \textit{dog \R{} man, dog \R{} woman, dog \R{} child, dog \R{} cat, dog \R{} rabbit, dog \R{} monkey, dog \R{} dog, dog \R{} hamster, dog \R{} kangaroo, dog \R{} panda,  }
    \item \textit{hamster \R{} man, hamster \R{} woman, hamster \R{} child, hamster \R{} cat, hamster \R{} rabbit, hamster \R{} monkey, hamster \R{} dog, hamster \R{} hamster, hamster \R{} kangaroo, hamster \R{} panda,}  
    \item \textit{kangaroo \R{} man, kangaroo \R{} woman, kangaroo \R{} child, kangaroo \R{} cat, kangaroo \R{} rabbit, kangaroo \R{} monkey, kangaroo \R{} dog, kangaroo \R{} hamster, kangaroo \R{} kangaroo, kangaroo \R{} panda,  }
    \item \textit{panda \R{} man, panda \R{} woman, panda \R{} child, panda \R{} cat, panda \R{} rabbit, panda \R{} monkey, panda \R{} dog, panda \R{} hamster, panda \R{} kangaroo, panda \R{} panda}
\end{itemize}

\section{Further Explanations on Basis Prepositions}
\label{sec:prepositions}

As stated in the manuscript, we devise a set of basis prepositions to steer the learning process of the relation prompt. Specifically, we collect a comprehensive list of $\sim$100 prepositions from~\cite{stevenson2010oxford}, and drop the prepositions that describes non-visual relations (\ie, temporal relations, causal relations, \textit{etc}.), while keep the ones that are related to visual relations. For example, the prepositional word \textit{``until"} is discarded as a temporal preposition, while words like \textit{``above"}, \textit{``beneath"}, \textit{``toward"} will be kept as plausible basis prepositions.

The basis preposition set contains a total of 56 words, listed in Table~\ref{tab:preposition_words}.

\section{Limitations}
\label{sec:limitations}

\begin{figure*}[t]
  \centering
   \includegraphics[width=0.99\linewidth]{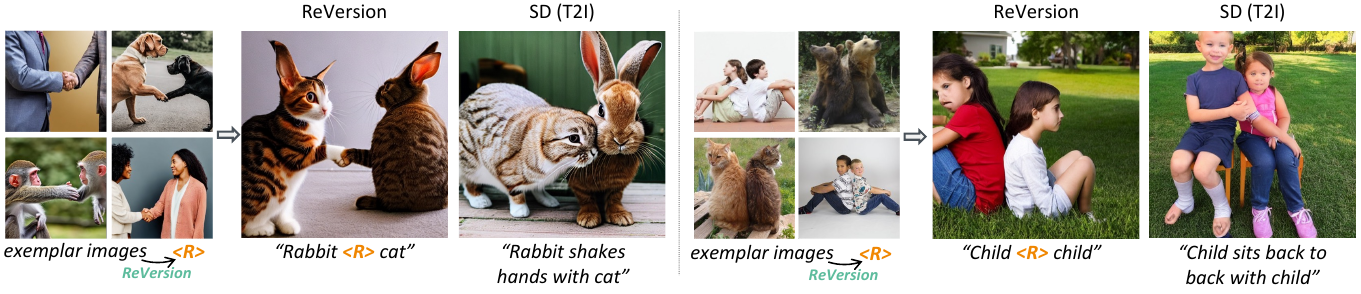}
   \vspace{-10pt}
   \caption{
        \textbf{Limitations}. Although the \R{} inverted by ReVersion can be applied robustly to synthesize new scenes, the image quality is limited by the generative capability of the pre-trained text-to-image model. \textit{\textbf{Left:}} when tasked with depicting a ``rabbit'' and a ``cat'' together, Stable Diffusion (SD) creates entities that blend features of both - such as rabbit ears and cat-like fur color and texture. Despite ReVersion's ability in capturing the ``shake hand'' relation through \R, the resulting image still has the problem of concept blending. \textbf{\textit{Right:}} when SD attempts to render human faces and bodies, the outcomes are often less than ideal. Therefore, even though ReVersion effectively captures the ``sitting back to back'' relation, the quality of the faces and bodies of the two children remains suboptimal.}

   \label{fig:fig_supp_limitation}
\end{figure*}

\begin{table}[t]
  \centering
  \caption{\textbf{Basis Preposition Set.} We list the set of 56 basis prepositions.}
  \vspace{-0.5em}
  \small 
  \begin{tabular}{|l|l|l|l|}
    \hline
    aboard & astride & in & regarding \\ \hline
    about & at & including & round \\ \hline
    above & atop & inside & through \\ \hline
    across & before & into & throughout \\ \hline
    after & behind & near & to \\ \hline
    against & below & of & toward \\ \hline
    along & beneath & off & towards \\ \hline
    alongside & beside & on & under \\ \hline
    amid & between & onto & underneath \\ \hline
    amidst & beyond & opposite & up \\ \hline
    among & by & out & upon \\ \hline
    amongst & down & outside & versus \\ \hline
    anti & following & over & with \\ \hline
    around & from & past & within \\ \hline
  \end{tabular}
  \label{tab:preposition_words}
\end{table}

Our performance is capped by the generative capabilities of the pre-trained text-to-image model, Stable Diffusion (SD). This dependency might lead to suboptimal synthesis in scenarios where SD faces challenges, as shown in Figure~\ref{fig:fig_supp_limitation}.

\noindent \textbf{Concept Blending.}
SD suffers from the concept blending problem. This issue arises when the model generates multiple entities within a single scene, leading to a fusion of characteristics from different classes. For example, when tasked with depicting a ``rabbit'' and a ``cat'' together, SD creates entities that blend features of both - such as rabbit ears and cat-like fur color and texture. Consequently, when ReVersion applies the learned \R{} on two entities of different classes, the same issue might occur.

\noindent \textbf{Human.} When SD attempts to render human faces and bodies, the outcomes are often less than ideal. Consequently, even though ReVersion effectively captures the relation, the quality of the faces and bodies of the human subjects might remain suboptimal.

Given that these limitations are inherent to the pre-trained text-to-image model, exploring and developing better text-to-image diffusion models is an orthogonal direction for performance improvements.

\section{Potential Societal Impacts}
\label{sec:impact}

Although \textit{ReVersion} can generate diverse entity combinations through inverted relations, this capability can also be exploited to synthesize real human figures interacting in ways they never did. As a result, we strongly advise users to only use \textit{ReVersion} for proper recreational purposes.

The rapid advancement of generative models has unlocked new levels of creativity but has also introduced various societal concerns. First, it is easier to create false imagery or manipulate data maliciously, leading to the spread of misinformation. Second, data used to train these models might be revealed during the sampling process without explicit consent from the data owner~\cite{tinsley2021face}. Third, generative models can suffer from the biases present in the training data~\cite{esser2020note}. We used the pre-trained Stable Diffusion~\cite{rombach2022ldm} for \textit{ReVersion}, which has been shown to suffer from data bias in certain scenarios. For example, when prompted with the phrase \textit{``a professor''}, Stable Diffusion tends to generate human figures that are white-passing and male-passing. We hope that more research will be conducted to address the risks and biases associated with generative models, and we advise everyone to use these models with discretion.

\section{More Qualitative Results}
\label{sec:qualitative}

We show various qualitative results in Figure~\ref{fig_supp_styles}-\ref{fig_supp_hanging_from_shake_hands}, which are located at the end of this Supplementary File.

\subsection{ReVersion with Diverse Styles and Backgrounds}
\begin{figure*}[t]
  \centering
  \vspace{30pt}
  \includegraphics[width=0.99\linewidth]{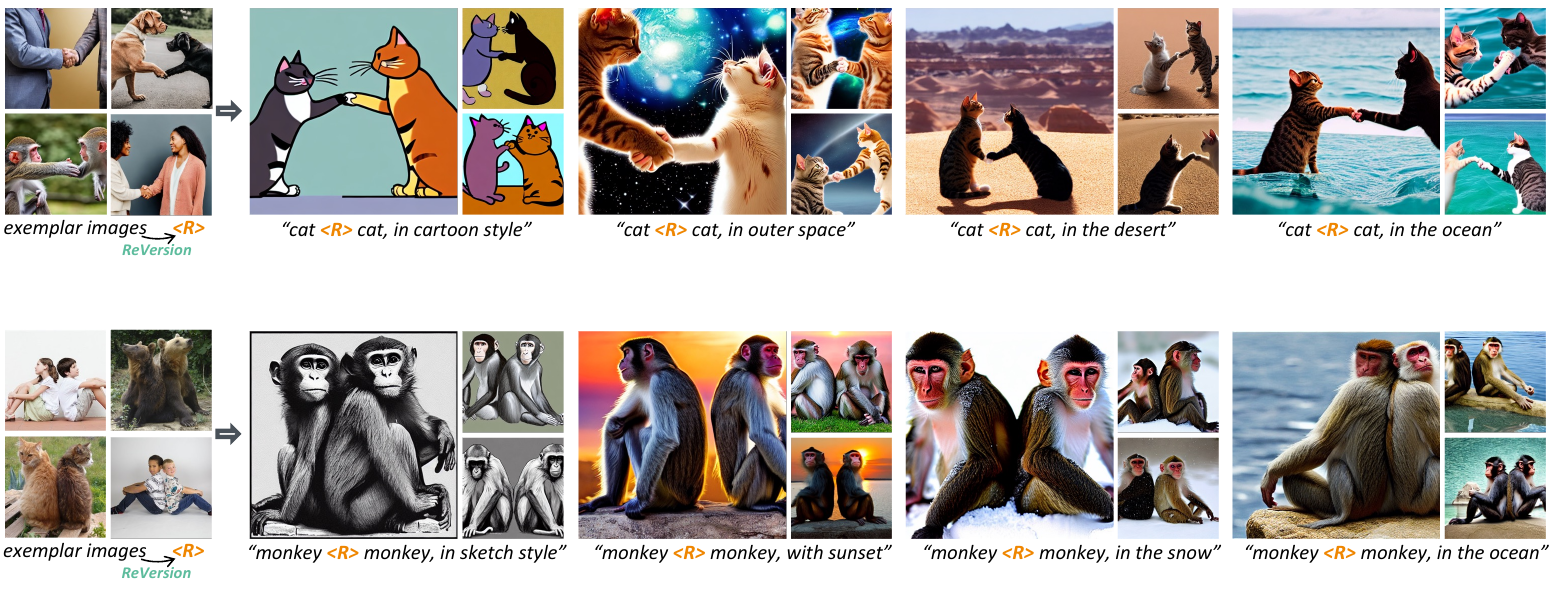}
   \caption{\textbf{ReVersion for Diverse Styles and Backgrounds}. The \R{} inverted by ReVersion can be applied robustly to relate entities in scenes with diverse backgrounds or styles.
   }
   \label{fig_supp_styles}
\end{figure*}

As shown in Figure~\ref{fig_supp_styles}, we apply the \R{} inverted by ReVersion in scenarios with diverse backgrounds and styles, and show that \R{} robustly adapt these environments with impressive results.

\subsection{ReVersion with Arbitrary Entity Combinations}
\begin{figure*}[t]
  \centering
   \includegraphics[width=0.6\linewidth]{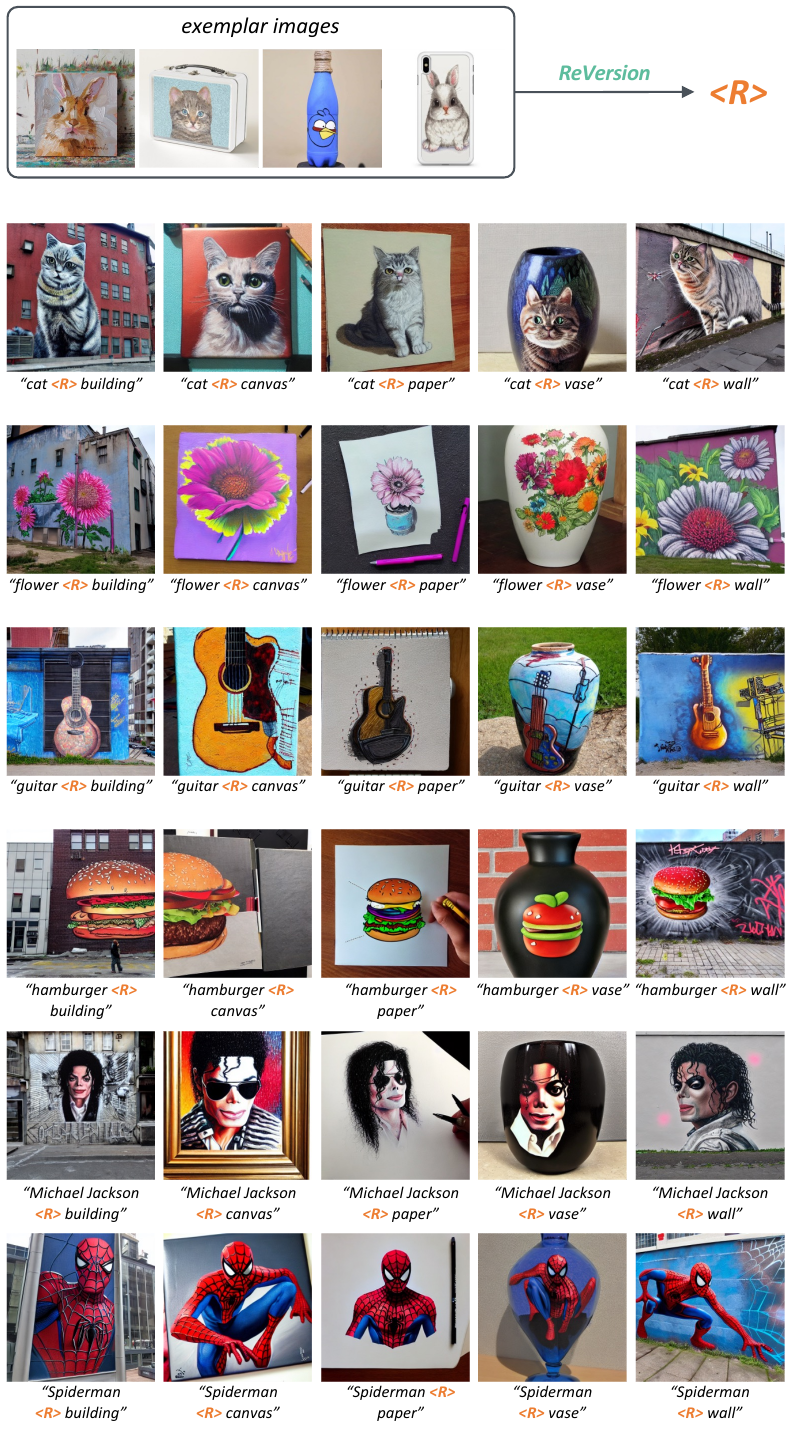}
   \caption{\textbf{Arbitrary Entity Combinations}. The \R{} inverted by ReVersion can be robustly applied to arbitrary entity combinations. For example, for the \R{}  extracted from the exemplar images where one entity is \textit{``painted on''} the other entity, we enumerate over all combinations among \textit{``\{cat / flower / guitar / hamburger / Michael Jackson / Spiderman\} \R{} \{building / canvas / paper / vase / wall\}''}, and observe that \R{} successfully links these entities together via exactly the same relation in the exemplar images.
   }
   \label{fig_supp_painted_on}
\end{figure*}

\begin{figure*}[t]
  \centering
   \includegraphics[width=0.95\linewidth]{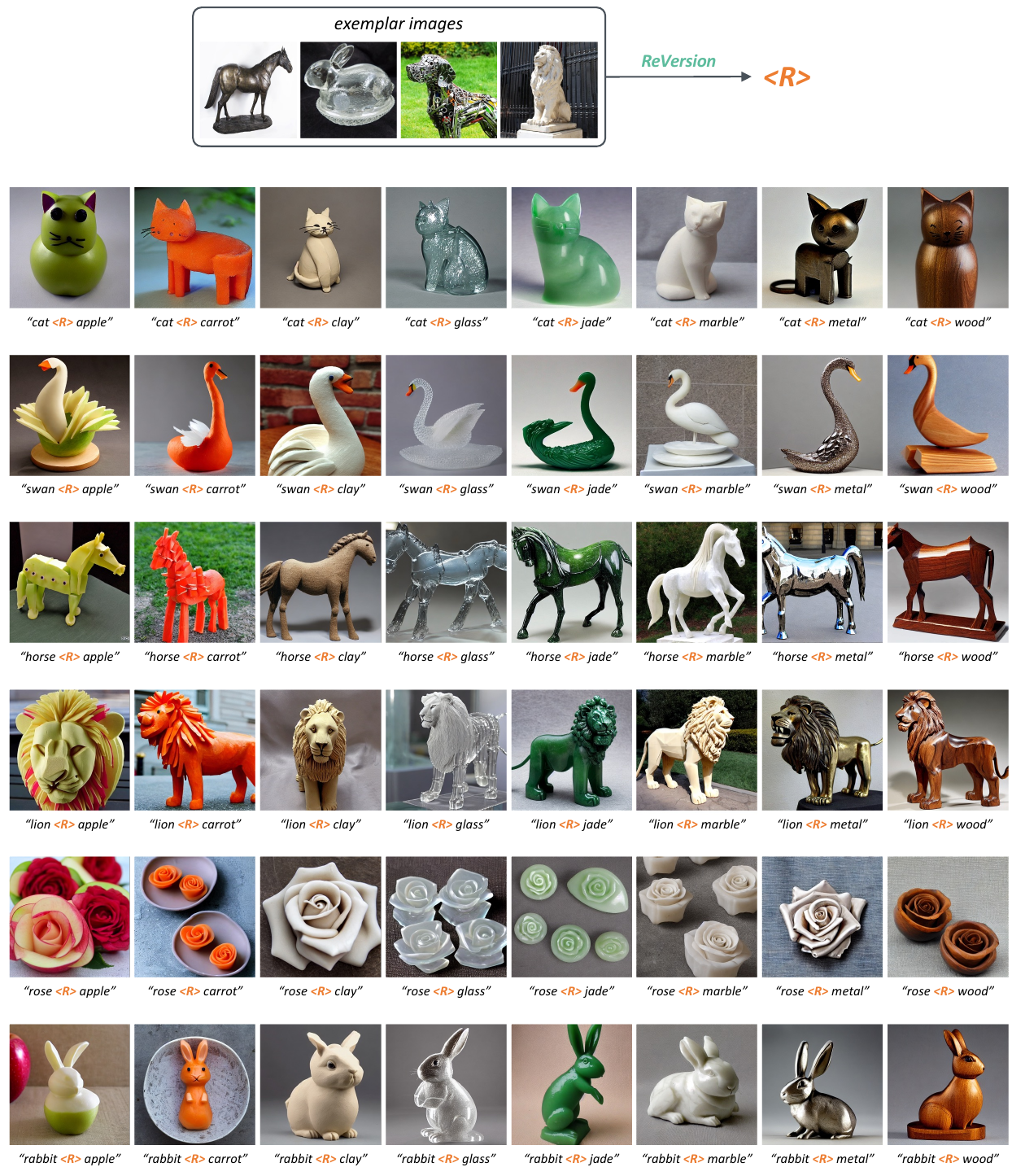}
   \caption{\textbf{Arbitrary Entity Combinations}. The \R{} inverted by ReVersion can be applied to arbitrary entity combinations. For example, for the \R{}  extracted from the exemplar images where one entity is \textit{``is made of the material of / is carved by''} the other entity, we enumerate over all combinations among \textit{``\{cat / swan / horse / lion / rose / rabbit\} \R{} \{apple / carrot / clay / glass / jade / marble / metal / wood\}''}, and observe that \R{} successfully links these entities together via exactly the same relation in the exemplar images.
   }
   \label{fig_supp_carved_by}
\end{figure*}

In Figure~\ref{fig_supp_painted_on} and \ref{fig_supp_carved_by}, we show that the \R{} inverted by ReVersion can be applied to robustly relate arbitrary entity combinations. For example, in Figure~\ref{fig_supp_painted_on}, for the \R{}  extracted from the exemplar images where one entity is \textit{``painted on''} the other entity, we enumerate over all combinations among \textit{``\{cat / flower / guitar / hamburger / Michael Jackson / Spiderman\} \R{} \{building / canvas / paper / vase / wall\}''}, and observe that \R{} successfully links these entities together via exactly the same relation in the exemplar images. 

\begin{figure*}[t]
  \centering
   \includegraphics[width=0.95\linewidth]{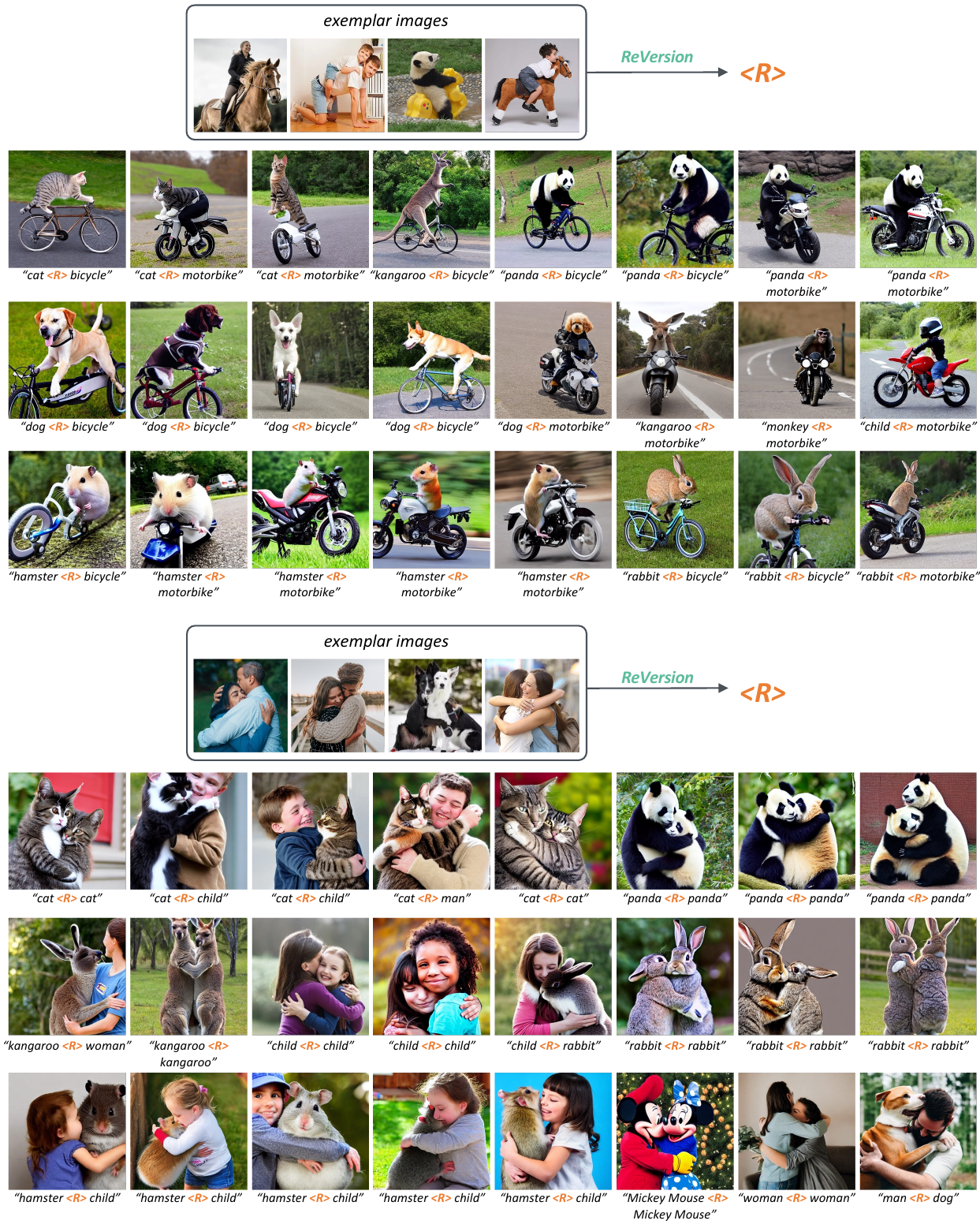}
   \caption{\textbf{More Qualitative Results}.
   }
   \label{fig_supp_ride_on_hug}
\end{figure*}

\begin{figure*}[t]
  \centering
   \includegraphics[width=0.99\linewidth]{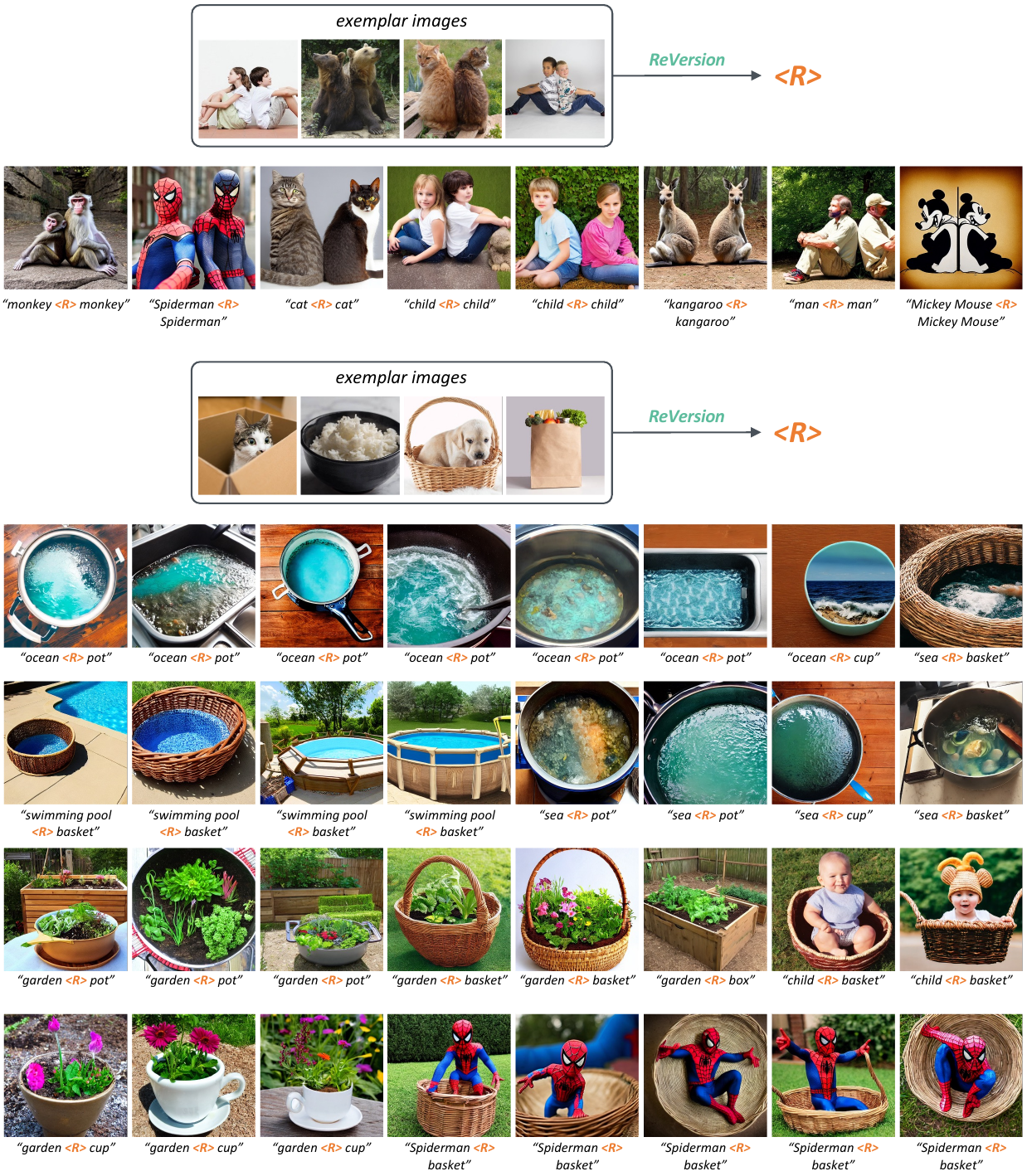}
   \caption{\textbf{More Qualitative Results}.
   }
   \label{fig_supp_inside_a_back2back}
\end{figure*}

\begin{figure*}[t]
  \centering
   \includegraphics[width=0.99\linewidth]{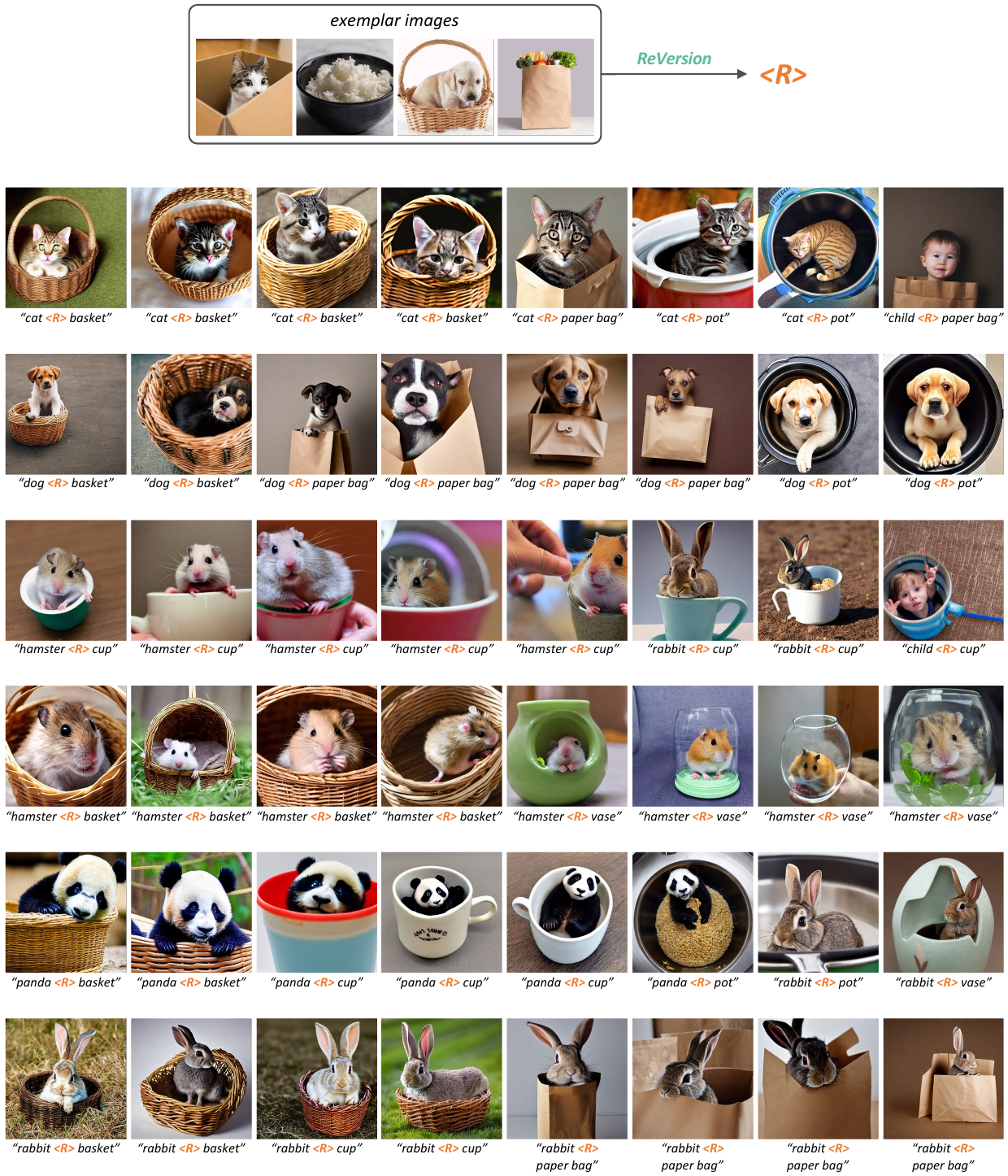}
   \caption{\textbf{More Qualitative Results}.
   }
   \label{fig_supp_inside_b}
\end{figure*}

\begin{figure*}[t]
  \centering
   \includegraphics[width=0.93\linewidth]{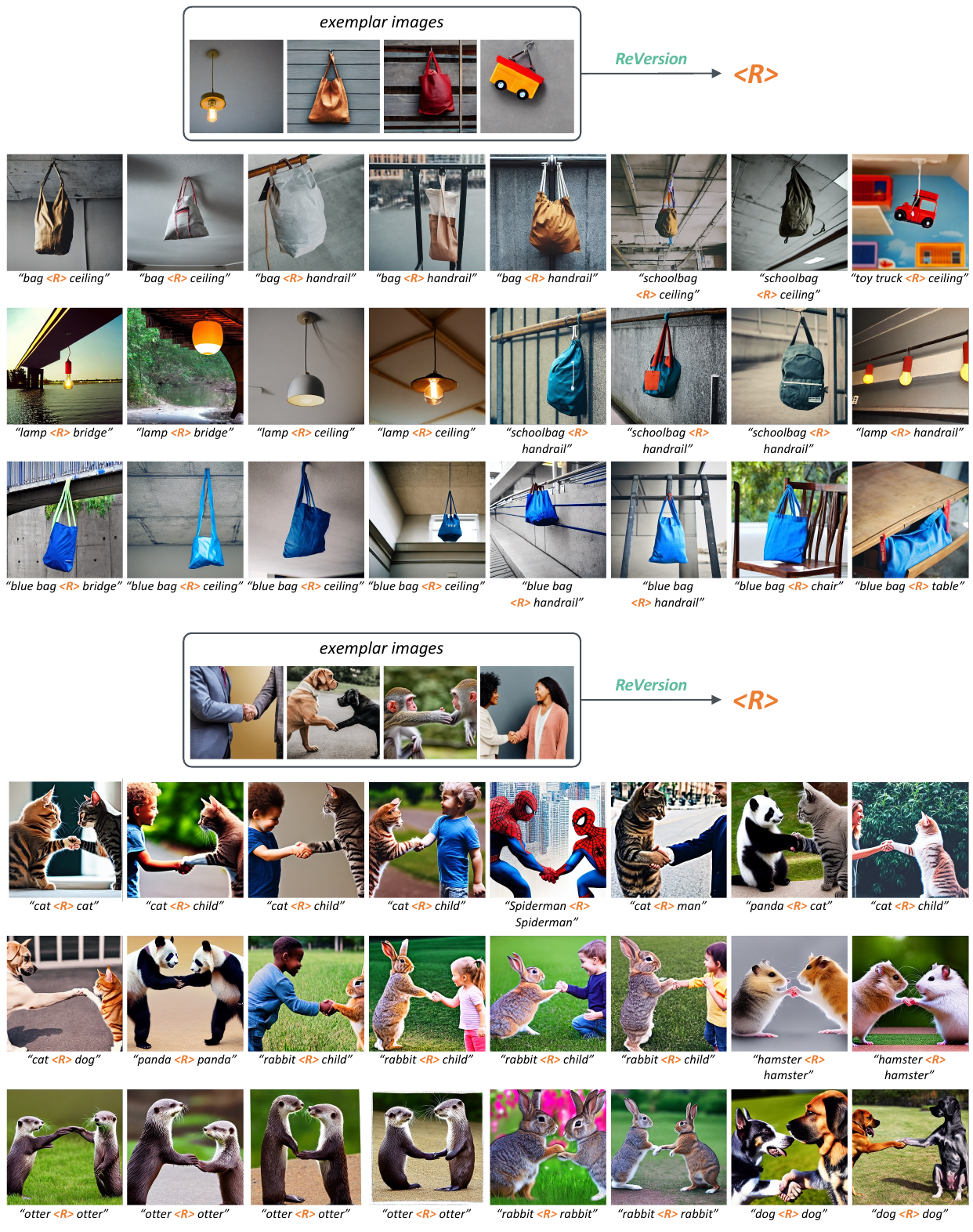}
   \caption{\textbf{More Qualitative Results}.
   }
   \label{fig_supp_hanging_from_shake_hands}
\end{figure*}

\subsection{Additional Qualitative Results}
We show additional qualitative results of ReVersion in Figure~\ref{fig_supp_ride_on_hug}, \ref{fig_supp_inside_a_back2back}, \ref{fig_supp_inside_b}, and \ref{fig_supp_hanging_from_shake_hands}.


\end{document}